\documentclass[runningheads]{llncs}
\usepackage{graphicx}
\usepackage{amsmath,amssymb} 
\usepackage{color}
\usepackage{dsfont}
\usepackage{multirow}
\usepackage{floatrow}
\usepackage{blindtext}
\usepackage[width=122mm,left=12mm,paperwidth=146mm,height=193mm,top=12mm,paperheight=217mm]{geometry}
\usepackage{epigraph}
\usepackage{booktabs}

\newcommand{\norm}[1]{\lVert#1\rVert}

\definecolor{MyDarkBlue}{rgb}{0,0.08,0.5}
\definecolor{MyLightBlue}{rgb}{0,0.08,0.8}
\definecolor{MyDarkGreen}{rgb}{0.02,0.50,0.02}
\definecolor{MyDarkRed}{rgb}{0.7,0.02,0.02}
\definecolor{MyDarkOrange}{rgb}{0.40,0.2,0.02}
\definecolor{MyDarkMagenta}{rgb}{0.337,0,0.827}

\begin{document}
\pagestyle{headings}
\mainmatter
\def\ECCV16SubNumber{}  

\title{Colorful Image Colorization} 

\titlerunning{Colorful Image Colorization}
\authorrunning{Zhang, Isola, Efros}

\author{Richard Zhang, Phillip Isola, Alexei A. Efros \\
\texttt{\{rich.zhang,isola,efros\}@eecs.berkeley.edu}}
\institute{University of California, Berkeley}

\maketitle

\begin{abstract}

Given a grayscale photograph as input, this paper attacks the problem of hallucinating a {\em plausible} color version of the photograph. This problem is clearly underconstrained, so previous approaches have either relied on significant user interaction or resulted in desaturated colorizations. We propose a fully automatic approach that produces vibrant and realistic colorizations. We embrace the underlying uncertainty of the problem by posing it as a classification task and use class-rebalancing at training time to increase the diversity of colors in the result. The system is implemented as a feed-forward pass in a CNN at test time and is trained on over a million color images. We evaluate our algorithm using a ``colorization Turing test," asking human participants to choose between a generated and ground truth color image. Our method successfully fools humans on 32\% of the trials, significantly higher than previous methods. Moreover, we show that colorization can be a powerful pretext task for self-supervised feature learning, acting as a {\em cross-channel encoder}. This approach results in state-of-the-art performance on several feature learning benchmarks.

\end{abstract}

\section{Introduction}

\begin{figure}[t]
 \centering
 \includegraphics[width=1.0\hsize]{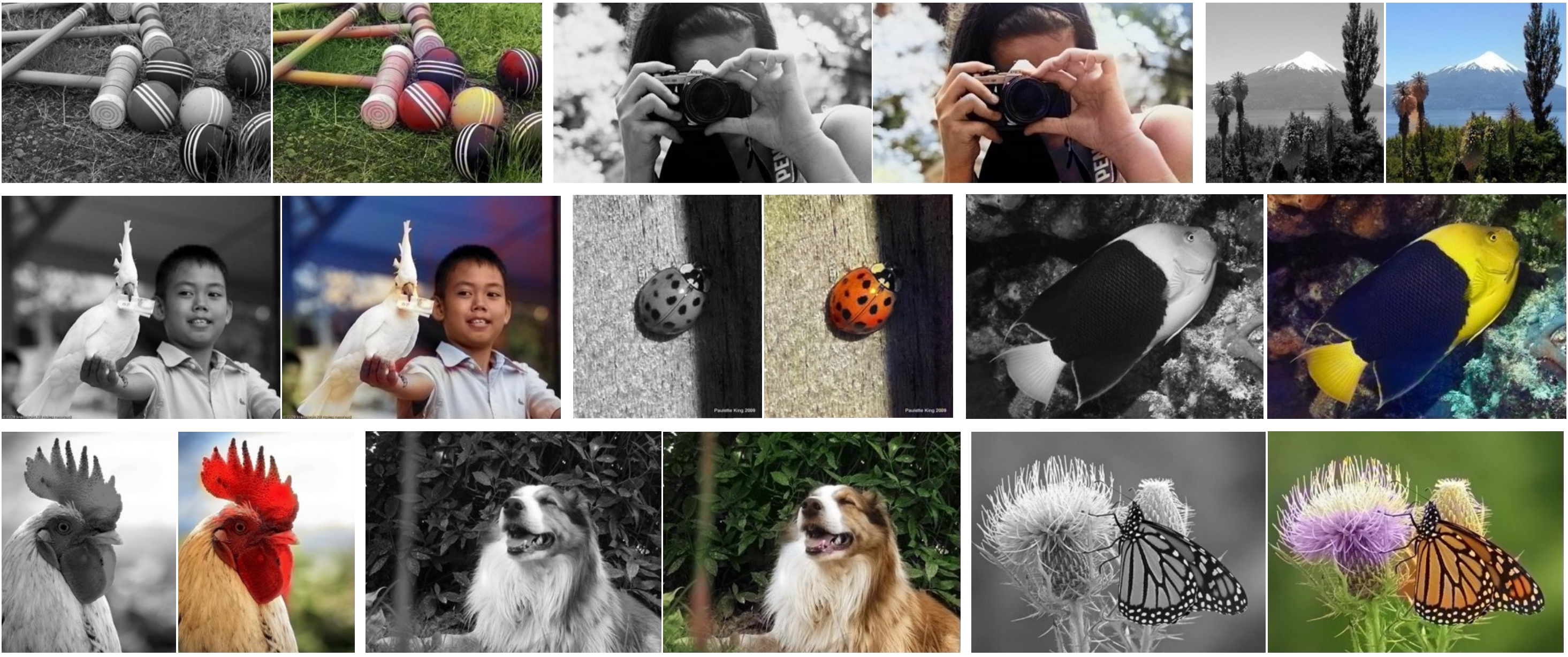}
  \caption{Example input grayscale photos and output colorizations from our algorithm. These examples are cases where our model works especially well. Please visit \texttt{http://richzhang.github.io/colorization/} to see the full range of results and to try our model and code. Best viewed in color (obviously).}
 \label{fig:teaser}
\end{figure}

Consider the grayscale photographs in Figure~\ref{fig:teaser}. At first glance, hallucinating their colors seems daunting, since so much of the information (two out of the three dimensions) has been lost. Looking more closely, however, one notices that in many cases, the semantics of the scene and its surface texture provide ample cues for many regions in each image: the grass is typically green, the sky is typically blue, and the ladybug is most definitely red. Of course, these kinds of semantic priors do not work for everything, e.g., the croquet balls on the grass might not, in reality, be red, yellow, and purple (though it's a pretty good guess). However, for this paper, our goal is not necessarily to recover the actual ground truth color, but rather to produce a {\em plausible} colorization that could potentially fool a human observer. Therefore, our task becomes much more achievable: to model enough of the statistical dependencies between the semantics and the textures of grayscale images and their color versions in order to produce visually compelling results.

Given the lightness channel $L$, our system predicts the corresponding $a$ and $b$ color channels of the image in the CIE $Lab$ colorspace. To solve this problem, we leverage large-scale data. Predicting color has the nice property that training data is practically free: any color photo can be used as a training example, simply by taking the image's $L$ channel as input and its $ab$ channels as the supervisory signal. Others have noted the easy availability of training data, and previous works have trained convolutional neural networks (CNNs) to predict color on large datasets \cite{cheng2015deep,dahl2016tinyclouds}. However, the results from these previous attempts tend to look desaturated. One explanation is that \cite{cheng2015deep,dahl2016tinyclouds} use loss functions that encourage conservative predictions. These losses are inherited from standard regression problems, where the goal is to minimize Euclidean error between an estimate and the ground truth.

We instead utilize a loss tailored to the colorization problem. As pointed out by \cite{charpiat2008automatic}, color prediction is inherently multimodal -- many objects can take on several plausible colorizations. For example, an apple is typically red, green, or yellow, but unlikely to be blue or orange. 
To appropriately model the multimodal nature of the problem, we predict a distribution of possible colors for each pixel. Furthermore, we re-weight the loss at training time to emphasize rare colors. This encourages our model to exploit the full diversity of the large-scale data on which it is trained. Lastly, we produce a final colorization by taking the \textit{annealed-mean} of the distribution. The end result is colorizations that are more vibrant and perceptually realistic than those of previous approaches. 



Evaluating synthesized images is notoriously difficult~\cite{ramanarayanan2007visual}. Since our ultimate goal is to make results that are compelling to a human observer, we introduce a novel way of evaluating colorization results, directly testing their perceptual realism. We set up a ``colorization Turing test," in which we show participants real and synthesized colors for an image, and ask them to identify the fake. In this quite difficult paradigm, we are able to fool participants on $32\%$ of the instances (ground truth colorizations would achieve 50\% on this metric), significantly higher than prior work \cite{dahl2016tinyclouds}. This test demonstrates that in many cases, our algorithm is producing nearly photorealistic results (see Figure \ref{fig:teaser} for selected successful examples from our algorithm). We also show that our system's colorizations are realistic enough to be useful for downstream tasks, in particular object classification, using an off-the-shelf VGG network \cite{simonyan2014very}.

We additionally explore colorization as a form of self-supervised representation learning, where raw data is used as its own source of supervision. The idea of learning feature representations in this way goes back at least to autoencoders \cite{bengio2013representation}. More recent works have explored feature learning via data imputation, where a held-out subset of the complete data is predicted (e.g., \cite{ngiam2011multimodal,agrawal2015learning,jayaraman2015learning,pathakCVPR16context,lotter2016deep,owens2016visually,owens2016ambient}). Our method follows in this line, and can be termed a \textit{cross-channel encoder}. We test how well our model performs in generalization tasks, compared to previous \cite{doersch2015unsupervised,agrawal2015learning,wang2015unsupervised,pathakCVPR16context} and concurrent \cite{donahue2016adversarial} self-supervision algorithms, and find that our method performs surprisingly well, achieving state-of-the-art performance on several metrics. 

Our contributions in this paper are in two areas. First, we make progress on the graphics problem of automatic image colorization by (a) designing an appropriate objective function that handles the multimodal uncertainty of the colorization problem and captures a wide diversity of colors, (b) introducing a novel framework for testing colorization algorithms, potentially applicable to other image synthesis tasks, and (c) setting a new high-water mark on the task by training on a million color photos. Secondly, we introduce the colorization task as a competitive and straightforward method for self-supervised representation learning, achieving state-of-the-art results on several benchmarks.

\textbf{Prior work on colorization} Colorization algorithms mostly differ in the ways they obtain and treat the data for modeling the correspondence between grayscale and color. Non-parametric methods, given an input grayscale image, first define one or more color reference images (provided by a user or retrieved automatically) to be used as source data.
Then, following the Image Analogies framework~\cite{hertzmann2001image}, color is transferred onto the input image from analogous regions of the reference image(s)~\cite{welsh2002transferring,gupta2012image,liu2008intrinsic,chia2011semantic}. Parametric methods, on the other hand, learn prediction functions from large datasets of color images at training time, posing the problem as either regression onto continuous color space~\cite{deshpande2015learning,cheng2015deep,dahl2016tinyclouds} or classification of quantized color values~\cite{charpiat2008automatic}. Our method also learns to classify colors, but does so with a larger model, trained on more data, and with several innovations in the loss function and mapping to a final continuous output.

\textbf{Concurrent work on colorization} Concurrently with our paper, Larsson et al. \cite{larsson2016learning} and Iizuka et al. \cite{iizuka2016let} have developed similar systems, which leverage large-scale data and CNNs. The methods differ in their CNN architectures and loss functions. While we use a classification loss, with rebalanced rare classes, Larsson et al. use an un-rebalanced classification loss, and Iizuka et al. use a regression loss. In Section \ref{sec:exp-imnet}, we compare the effect of each of these types of loss function in conjunction with our architecture. The CNN architectures are also somewhat different: Larsson et al. use hypercolumns \cite{hariharan2015hypercolumns} on a VGG network \cite{simonyan2014very}, Iizuka et al. use a two-stream architecture in which they fuse global and local features, and we use a single-stream, VGG-styled network with added depth and dilated convolutions \cite{chen2016deeplab,yu2015multi}. In addition, while we and Larsson et al. train our models on ImageNet \cite{russakovsky2015imagenet}, Iizuka et al. train their model on Places \cite{zhou2014learning}. In Section \ref{sec:exp-imnet}, we provide quantitative comparisons to Larsson et al., and encourage interested readers to investigate both concurrent papers.

\begin{figure}[t]
 \centering
 \includegraphics[width=1.0\hsize]{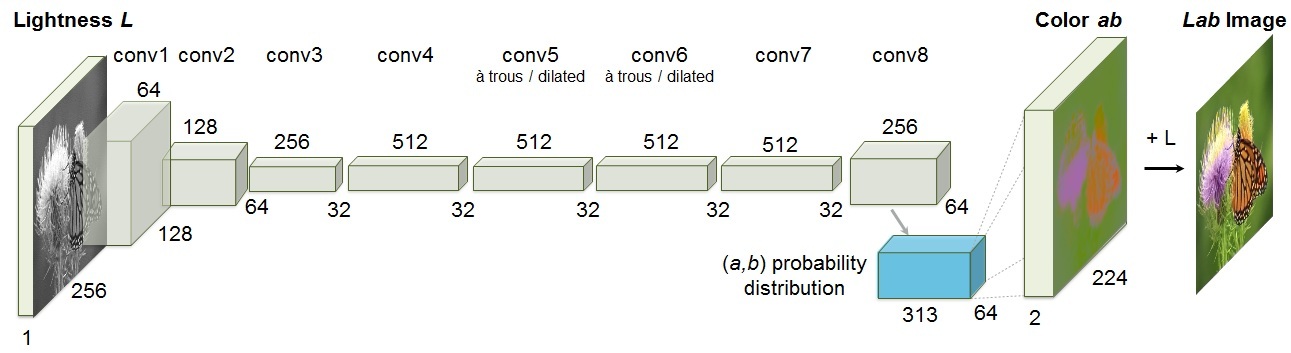}
  \caption{Our network architecture. Each \texttt{conv} layer refers to a block of 2 or 3 repeated \texttt{conv} and \texttt{ReLU} layers, followed by a \texttt{BatchNorm} \cite{ioffe2015batch} layer. The net has no \texttt{pool} layers. All changes in resolution are achieved through spatial downsampling or upsampling between \texttt{conv} blocks.}
 \label{fig:architecture}
\end{figure}

\section{Approach}

We train a CNN to map from a grayscale input to a distribution over quantized color value outputs using the architecture shown in Figure \ref{fig:architecture}. Architectural details are described in the supplementary materials on our project webpage\footnote{\texttt{http://richzhang.github.io/colorization/}}, and the model is publicly available. In the following, we focus on the design of the objective function, and our technique for inferring point estimates of color from the predicted color distribution.

\subsection{Objective Function}
\label{sec:obj}
Given an input lightness channel $\mathbf{X}\in\mathds{R}^{H\times W\times1}$, our objective is to learn a mapping $\mathbf{\widehat{Y}} = \mathcal{F}(\mathbf{X})$ to the two associated color channels $\mathbf{Y} \in \mathds{R}^{H\times W \times2}$, where $H,W$ are image dimensions.

(We denote predictions with a $\widehat{\cdot}$ symbol and ground truth without.) We perform this task in CIE \textit{Lab} color space. Because distances in this space model perceptual distance, a natural objective function, as used in \cite{cheng2015deep,dahl2016tinyclouds}, is the Euclidean loss $\text{L}_{2}(\cdot,\cdot)$ between predicted and ground truth colors:

\begin{equation}
\text{L}_{2}(\mathbf{\widehat{Y}},\mathbf{Y}) = \frac{1}{2} \sum_{h,w} \norm{\mathbf{Y}_{h,w}-\mathbf{\widehat{Y}}_{h,w}}_2^2
\label{eqn:top_L2}
\end{equation}

However, this loss is not robust to the inherent ambiguity and multimodal nature of the colorization problem. If an object can take on a set of distinct \textit{ab} values, the optimal solution to the Euclidean loss will be the mean of the set. In color prediction, this averaging effect favors grayish, desaturated results. Additionally, if the set of plausible colorizations is non-convex, the solution will in fact be out of the set, giving implausible results.

\begin{figure}[t]
 \centering
 \includegraphics[width=1.0\hsize]{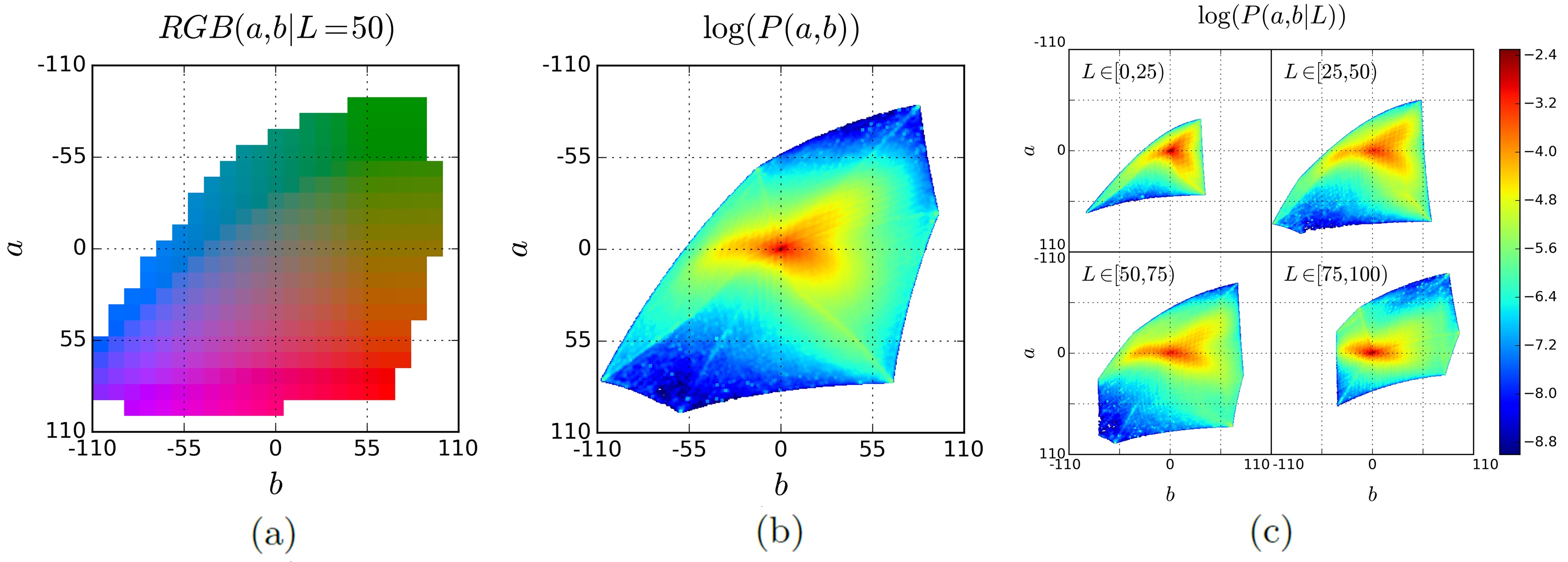}
 \caption{(a) Quantized \textit{ab} color space with a grid size of 10. A total of 313 \textit{ab} pairs are in gamut. (b) Empirical probability distribution of $ab$ values, shown in log scale. (c) Empirical probability distribution of $ab$ values, conditioned on \textit{L}, shown in log scale.}
 \label{fig:ab_space}
\end{figure}

Instead, we treat the problem as multinomial classification. We quantize the \textit{ab} output space into bins with grid size 10 and keep the $Q=313$ values which are in-gamut, as shown in Figure \ref{fig:ab_space}(a). For a given input $\mathbf{X}$, we learn a mapping $\mathbf{\widehat{Z}} = \mathcal{G}(\mathbf{X})$ to a probability distribution over possible colors $\mathbf{\widehat{Z}}\in [0,1]^{H\times W\times Q}$, where $Q$ is the number of quantized \textit{ab} values.


To compare predicted $\mathbf{\widehat{Z}}$ against ground truth, we define function $\mathbf{Z} = \mathcal{H}_{gt}^{-1}(\mathbf{Y})$, which converts ground truth color $\mathbf{Y}$ to vector $\mathbf{Z}$, using a soft-encoding scheme\footnote{Each ground truth value $\mathbf{Y}_{h,w}$ can be encoded as a 1-hot vector $\mathbf{Z}_{h,w}$ by searching for the nearest quantized $ab$ bin. However, we found that \textit{soft}-encoding worked well for training, and allowed the network to quickly learn the relationship between elements in the output space \cite{hinton2015distilling}. We find the 5-nearest neighbors to $\mathbf{Y}_{h,w}$ in the output space and weight them proportionally to their distance from the ground truth using a Gaussian kernel with $\sigma=5$.}. We then use multinomial cross entropy loss $\text{L}_{cl}(\cdot,\cdot)$, defined as:

\begin{equation}
\text{L}_{cl}(\mathbf{\widehat{Z}},\mathbf{Z}) = - \sum_{h,w} v(\mathbf{Z}_{h,w}) \sum_{q} \mathbf{Z}_{h,w,q} \log (\mathbf{\widehat{Z}}_{h,w,q})
\label{eqn:multiloss}
\end{equation}

where $v(\cdot)$ is a weighting term that can be used to rebalance the loss based on color-class rarity, as defined in Section \ref{sec:class_rebal} below. Finally, we map probability distribution $\mathbf{\widehat{Z}}$ to color values $\mathbf{\widehat{Y}}$ with function $\mathbf{\widehat{Y}} = \mathcal{H}(\mathbf{\widehat{Z}})$, which will be further discussed in Section \ref{sec:final_pred}.



\subsection{Class rebalancing}
\label{sec:class_rebal}

The distribution of \textit{ab} values in natural images is strongly biased towards values with low $ab$ values, due to the appearance of backgrounds such as clouds, pavement, dirt, and walls. Figure \ref{fig:ab_space}(b) shows the empirical distribution of pixels in \textit{ab} space, gathered from 1.3M training images in ImageNet \cite{russakovsky2015imagenet}. Observe that the number of pixels in natural images at desaturated values are orders of magnitude higher than for saturated values. Without accounting for this, the loss function is dominated by desaturated \textit{ab} values. 
We account for the class-imbalance problem by reweighting the loss of each pixel at train time based on the pixel color rarity. This is asymptotically equivalent to the typical approach of resampling the training space \cite{farabet2013learning}. Each pixel is weighed by factor $\mathbf{w} \in \mathds{R}^{Q}$, based on its closest $ab$ bin.

\begin{gather}
v(\mathbf{Z}_{h,w}) = \mathbf{w}_{q^{*}}, \text{ where } q^{*} = \arg\max_{q} \mathbf{Z}_{h,w,q}
\label{eqn:rebal_v} \\
\mathbf{w} \propto \Big( (1-\lambda) \mathbf{\widetilde{p}} + \frac{\lambda}{Q} \Big)^{-1}, \hspace{.15in}
\mathds{E} [\mathbf{w}] = \sum_{q} \mathbf{\widetilde{p}}_q \mathbf{w}_q = 1
\label{eqn:rebal_w}
\end{gather}

To obtain smoothed empirical distribution $\mathbf{\widetilde{p}} \in \Delta^Q$, we estimate the empirical probability of colors in the quantized $ab$ space $\mathbf{p} \in \Delta^Q$ from the full ImageNet training set and smooth the distribution with a Gaussian kernel $\mathbf{G}_\sigma$. We then mix the distribution with a uniform distribution with weight $\lambda\in [0,1]$, take the reciprocal, and normalize so the weighting factor is 1 on expectation. We found that values of $\lambda=\tfrac{1}{2}$ and $\sigma=5$ worked well. We compare results with and without class rebalancing in Section \ref{sec:comp_imnet}.

\subsection{Class Probabilities to Point Estimates}
\label{sec:final_pred}
\begin{figure}[t]
\centering
\includegraphics[width=1.\hsize]{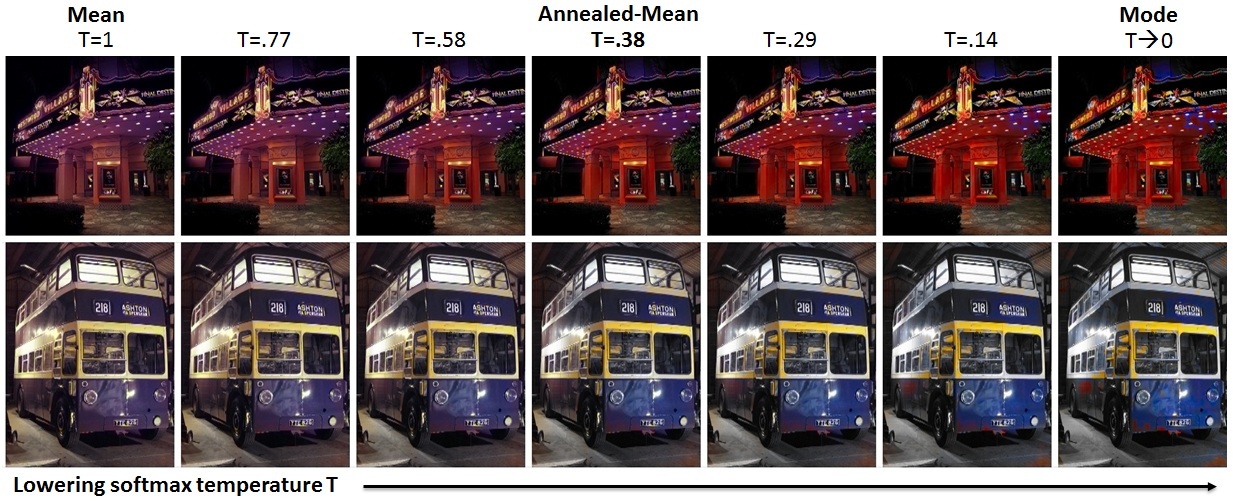}
\caption{The effect of temperature parameter $T$ on the \textit{annealed-mean} output (Equation \ref{eqn:ann-mean}).
The left-most images show the means of the predicted color distributions and the right-most show the modes. We use $T=0.38$ in our system.}
\label{fig:temp_sweep}
\end{figure}

Finally, we define $\mathcal{H}$, which maps the predicted distribution $\mathbf{\widehat{Z}}$ to point estimate $\mathbf{\widehat{Y}}$ in \textit{ab} space. One choice is to take the mode of the predicted distribution for each pixel, as shown in the right-most column of Figure~\ref{fig:temp_sweep} for two example images. This provides a vibrant but sometimes spatially inconsistent result, e.g., the red splotches on the bus. 
On the other hand, taking the mean of the predicted distribution produces spatially consistent but desaturated results (left-most column of Figure~\ref{fig:temp_sweep}), exhibiting an unnatural sepia tone. 
This is unsurprising, as taking the mean after performing classification suffers from some of the same issues as optimizing for a Euclidean loss in a regression framework. To try to get the best of both worlds, we \textit{interpolate} by re-adjusting the temperature $T$ of the softmax distribution, and taking the mean of the result. We draw inspiration from the simulated annealing technique \cite{kirkpatrick1983optimization}, and thus refer to the operation as taking the \textit{annealed-mean} of the distribution:

\begin{equation}
\mathcal{H}(\mathbf{Z}_{h,w}) = \mathds{E}\big[ f_{T}(\mathbf{Z}_{h,w}) \big], \hspace{0.15in}
f_{T}(\mathbf{z}) = \dfrac{\exp(\log(\mathbf{z})/T)}{\sum_{q} \exp(\log(\mathbf{z}_q)/T)}
\label{eqn:ann-mean}
\end{equation}

Setting $T=1$ leaves the distribution unchanged, lowering the temperature $T$ produces a more strongly peaked distribution, and setting $T\rightarrow 0$ results in a 1-hot encoding at the distribution mode. We found that temperature $T=0.38$, shown in the middle column of Figure \ref{fig:temp_sweep}, captures the vibrancy of the mode while maintaining the spatial coherence of the mean. 

Our final system $\mathcal{F}$ is the composition of CNN $\mathcal{G}$, which produces a predicted distribution over all pixels, and the annealed-mean operation $\mathcal{H}$, which produces a final prediction. The system is not quite end-to-end trainable, but note that the mapping $\mathcal{H}$ operates on each pixel independently, with a single parameter, and can be implemented as part of a feed-forward pass of the CNN.

\section{Experiments}

In Section \ref{sec:exp-imnet}, we assess the graphics aspect of our algorithm, evaluating the perceptual realism of our colorizations, along with other measures of accuracy. We compare our full algorithm to several variants, along with recent~\cite{dahl2016tinyclouds} and concurrent work~\cite{larsson2016learning}. In Section~\ref{sec:exp-feat}, we test colorization as a method for self-supervised representation learning. Finally, in Section~\ref{sec:comp_heldout}, we show qualitative examples on legacy black and white images. 

\subsection{Evaluating colorization quality}
\label{sec:exp-imnet}

\begin{table}[t]
\centering
\scalebox{0.9} {
\begin{tabular}{lccccccc}
\specialrule{.1em}{.1em}{.1em}
\multicolumn{8}{c}{\textbf{Colorization Results on ImageNet}} \\
\specialrule{.1em}{.1em}{.1em}
\specialrule{.1em}{.1em}{.1em}
\multicolumn{1}{c}{} & \multicolumn{3}{c}{\textbf{Model}} & \multicolumn{2}{c}{\textbf{AuC}} & \textbf{VGG Top-1} & \textbf{AMT} \\
\textbf{Method} & \textbf{Params} & \textbf{Feats} & \multicolumn{1}{c}{\textbf{Runtime}} & \textbf{non-rebal} & \textbf{rebal} & \textbf{Class Acc} & \textbf{Labeled} \\
\multicolumn{1}{c}{\textbf{}} & \textbf{(MB)} & \textbf{(MB)} & \multicolumn{1}{c}{\textbf{(ms)}} & \textbf{(\%)} & \textbf{(\%)} & \textbf{(\%)} & \textbf{Real (\%)} \\ \hline
Ground Truth & -- & -- & -- & 100 & 100 & 68.3 & 50 \\
Gray & -- & -- & -- & 89.1 & 58.0 & 52.7 & -- \\
Random & -- & -- & -- & 84.2 & 57.3 & 41.0 & 13.0$\pm$4.4 \\ \hline
Dahl \cite{dahl2016tinyclouds} & -- & -- & -- & 90.4 & 58.9 & 48.7 & 18.3$\pm$2.8 \\
Larsson et al. \cite{larsson2016learning} & 588 & 495 & 122.1 & \textbf{91.7} & 65.9 & \textbf{59.4} & \textbf{27.2}$\pm$\textbf{2.7} \\ \hline
Ours (L2) & 129 & 127 & 17.8 & 91.2 & 64.4 & 54.9 & 21.2$\pm$2.5 \\
Ours (L2, ft) & 129 & 127 & 17.8 & 91.5 & 66.2 & 56.5 & 23.9$\pm$2.8 \\
Ours (class) & 129 & 142 & 22.1 & 91.6 & 65.1 & 56.6 & 25.2$\pm$2.7 \\ \hline
Ours (full) & 129 & 142 & 22.1 & 89.5 & \textbf{67.3} & 56.0 & \textbf{32.3}$\pm$\textbf{2.2} \\ \specialrule{.1em}{.1em}{.1em}
\label{tab:res_imnet}
\end{tabular}
}
\caption{Colorization results on 10k images in the ImageNet validation set \cite{russakovsky2015imagenet}, as used in \cite{larsson2016learning}. AuC refers to the area under the curve of the cumulative error distribution over \textit{ab} space~\cite{deshpande2015learning}. Results column 2 shows the class-balanced variant of this metric. Column 3 is the classification accuracy after colorization using the VGG-16 \cite{simonyan2014very} network. Column 4 shows results from our AMT \textit{real vs. fake} test (with mean and standard error reported, estimated by bootstrap~\cite{efron1992bootstrap}). Note that an algorithm that produces ground truth images would achieve 50\% performance in expectation. Higher is better for all metrics. Rows refer to different algorithms; see text for a description of each. Parameter and feature memory, and runtime, were measured on a Titan X GPU using \textit{Caffe} \cite{jia2014caffe}.}
\end{table}

\begin{figure}
 \centering
 \includegraphics[width=1\hsize]{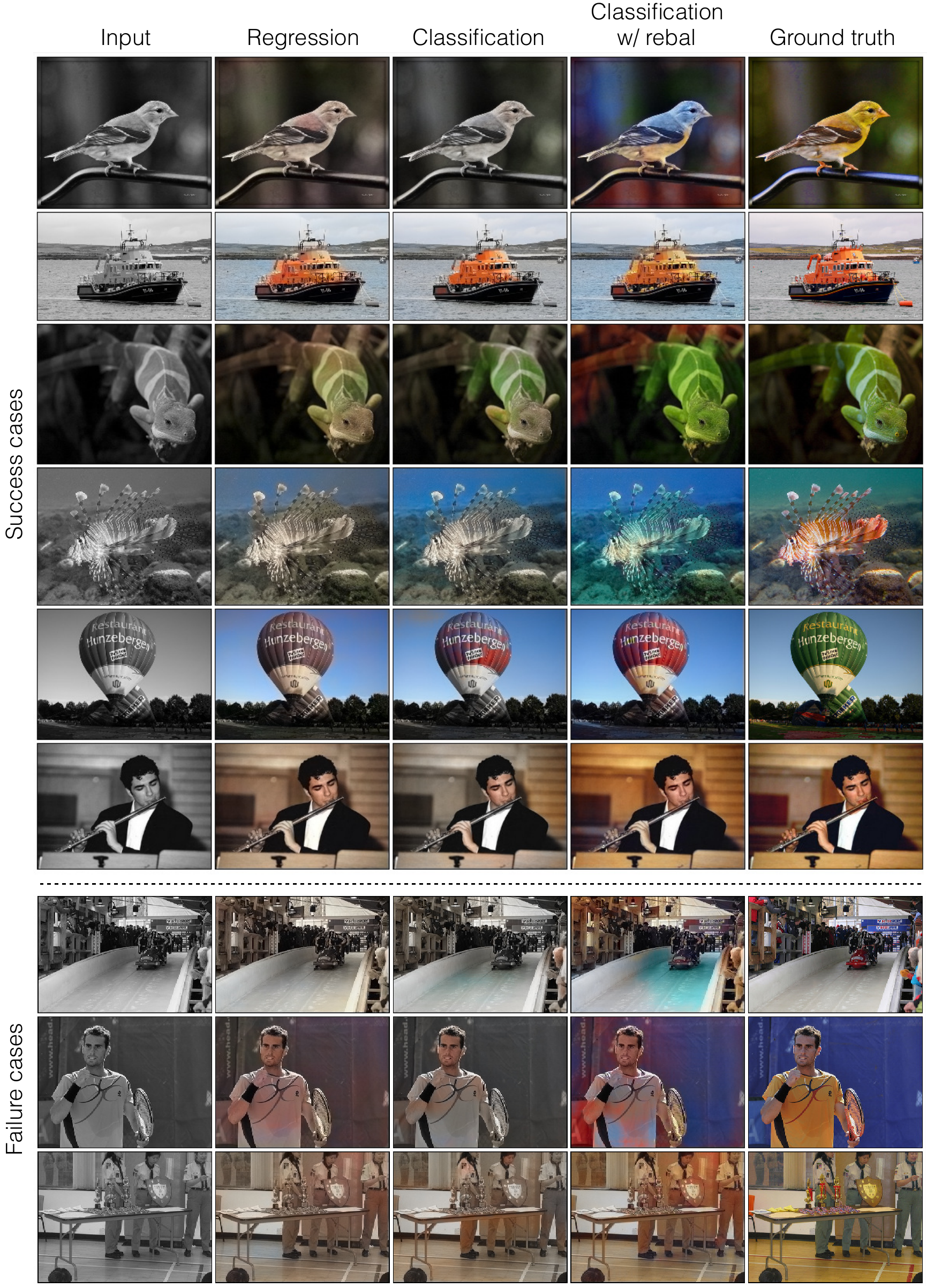}
   \caption{Example results from our ImageNet test set. Our classification loss with rebalancing produces more accurate and vibrant results than a regression loss or a classification loss without rebalancing. Successful colorizations are above the dotted line. Common failures are below. These include failure to capture long-range consistency, frequent confusions between red and blue, and a default sepia tone on complex indoor scenes. Please visit \texttt{http://richzhang.github.io/colorization/} to see the full range of results.}
  \label{fig:imagenet_comparison}
\end{figure}

\label{sec:comp_imnet}
We train our network on the 1.3M images from the ImageNet training set~\cite{russakovsky2015imagenet}, validate on the first 10k images in the ImageNet validation set, and test on a separate 10k images in the validation set, same as in~\cite{larsson2016learning}.
We show quantitative results in Table 1 on three metrics. A qualitative comparison for selected success and failure cases is shown in Figure~\ref{fig:imagenet_comparison}. For a comparison on a full selection of random images, please see our project webpage.


To specifically test the effect of different loss functions, we train our CNN with various losses. We also compare to previous~\cite{dahl2016tinyclouds} and concurrent methods~\cite{larsson2016learning}, which both use CNNs trained on ImageNet, along with naive baselines:
\begin{enumerate}
\item \textbf{Ours (full)} Our full method, with classification loss, defined in Equation~\ref{eqn:multiloss}, and class rebalancing, as described in Section~\ref{sec:class_rebal}. The network was trained from scratch with k-means initialization~\cite{krahenbuhl2015data}, using the ADAM solver for approximately 450k iterations\footnote{$\beta_1=.9$, $\beta_2=.99$, and weight decay = $10^{-3}$. Initial learning rate was $3\times10^{-5}$ and dropped to $10^{-5}$ and $3\times10^{-6}$ when loss plateaued, at 200k and 375k iterations, respectively. Other models trained from scratch followed similar training protocol.}.
\item \textbf{Ours (class)} Our network on classification loss but no class rebalancing ($\lambda=1$ in Equation \ref{eqn:rebal_w}).
\item \textbf{Ours (L2)} Our network trained from scratch, with L2 regression loss, described in Equation~\ref{eqn:top_L2}, following the same training protocol.
\item \textbf{Ours (L2, ft)} Our network trained with L2 regression loss, fine-tuned from our full classification with rebalancing network.
\item \textbf{Larsson et al.~\cite{larsson2016learning}} A CNN method that also appears in these proceedings.
\item \textbf{Dahl \cite{dahl2016tinyclouds}} A previous model using a Laplacian pyramid on VGG features, trained with L2 regression loss.
\item \textbf{Gray} Colors every pixel gray, with $(a,b)=0$.
\item \textbf{Random} Copies the colors from a random image from the training set. 
\end{enumerate}

Evaluating the quality of synthesized images is well-known to be a difficult task, as simple quantitative metrics, like RMS error on pixel values, often fail to capture visual realism. To address the shortcomings of any individual evaluation, we test three that measure different senses of quality, shown in Table 1.

\textbf{1. Perceptual realism (AMT):} For many applications, such as those in graphics, the ultimate test of colorization is how compelling the colors look to a human observer. To test this, we ran a \textit{real vs. fake} two-alternative forced choice experiment on Amazon Mechanical Turk (AMT). Participants in the experiment were shown a series of pairs of images. Each pair consisted of a color photo next to a re-colorized version, produced by either our algorithm or a baseline. Participants were asked to click on the photo they believed contained \textit{fake} colors generated by a computer program. Individual images of resolution $256\times256$ were shown for one second each, and after each pair, participants were given unlimited time to respond. Each experimental session consisted of 10 practice trials (excluded from subsequent analysis), followed by 40 test pairs. On the practice trials, participants were given feedback as to whether or not their answer was correct. No feedback was given during the 40 test pairs. Each session tested only a single algorithm at a time, and participants were only allowed to complete at most one session. A total of 40 participants evaluated each algorithm. To ensure that all algorithms were tested in equivalent conditions (i.e. time of day, demographics, etc.), all experiment sessions were posted simultaneously and distributed to Turkers in an i.i.d. fashion.


\begin{figure}[t]
 \centering
 \includegraphics[width=1.0\hsize]{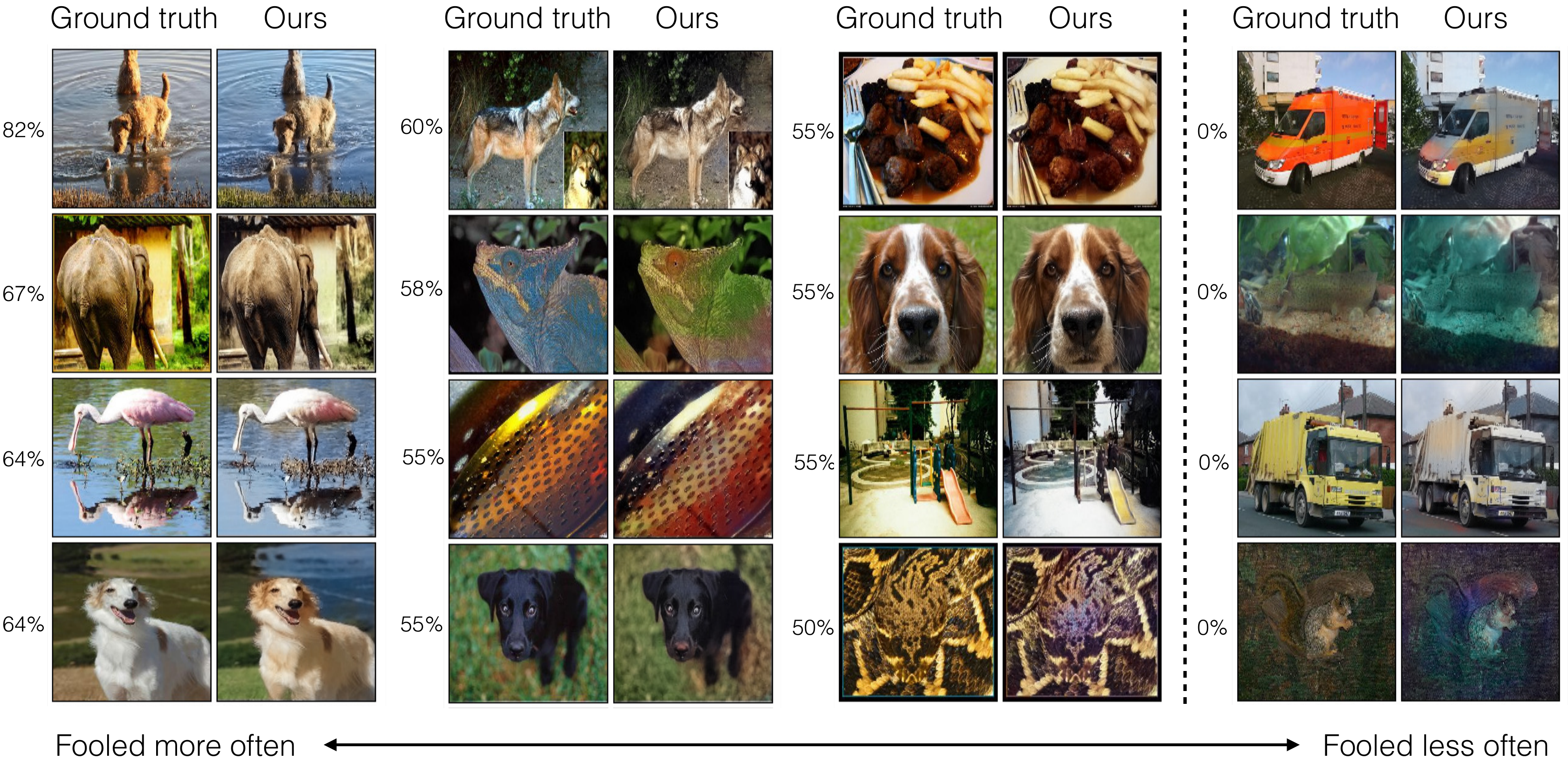}
  \caption{Images sorted by how often AMT participants chose our algorithm's colorization over the ground truth. In all pairs to the left of the dotted line, participants believed our colorizations to be more real than the ground truth on $\geq 50\%$ of the trials. In some cases, this may be due to poor white balancing in the ground truth image, corrected by our algorithm, which predicts a more prototypical appearance. Right of the dotted line are examples where participants were never fooled.}
 \label{fig:top_bottom_turk}
\end{figure}

To check that participants were competent at this task, $10\%$ of the trials pitted the ground truth image against the Random baseline described above. Participants successfully identified these random colorizations as fake $87\%$ of the time, indicating that they understood the task and were paying attention. 

Figure \ref{fig:top_bottom_turk} gives a better sense of the participants' competency at detecting subtle errors made by our algorithm. The far right column shows example pairs where participants identified the fake image successfully in $100\%$ of the trials. Each of these pairs was scored by at least 10 participants. Close inspection reveals that on these images, our colorizations tend to have giveaway artifacts, such as the yellow blotches on the two trucks, which ruin otherwise decent results.

Nonetheless, our full algorithm fooled participants on $32\%$ of trials, as shown in Table 1. This number is significantly higher than all compared algorithms ($p < 0.05$ in each case) except for Larsson et al., against which the difference was not significant ($p = 0.10$; all statistics estimated by bootstrap \cite{efron1992bootstrap}). These results validate the effectiveness of using both a classification loss and class-rebalancing.


Note that if our algorithm exactly reproduced the ground truth colors, the forced choice would be between two identical images, and participants would be fooled $50\%$ of the time on expectation. Interestingly, we can identify cases where participants were fooled \emph{more} often than $50\%$ of the time, indicating our results were deemed more realistic than the ground truth. Some examples are shown in the first three columns of Figure \ref{fig:top_bottom_turk}. In many case, the ground truth image is poorly white balanced or has unusual colors, whereas our system produces a more prototypical appearance.

\textbf{2. Semantic interpretability (VGG classification):} Does our method produce realistic enough colorizations to be interpretable to an off-the-shelf object classifier? We tested this by feeding our \textit{fake} colorized images to a VGG network \cite{simonyan2014very} that was trained to predict ImageNet classes from \textit{real} color photos. If the classifier performs well, that means the colorizations are accurate enough to be informative about object class. Using an off-the-shelf classifier to assess the realism of synthesized data has been previously suggested by \cite{owens2016visually}.

The results are shown in the second column from the right of Table 1. Classifier performance drops from 68.3\% to 52.7\% after ablating colors from the input. After re-colorizing using our full method, the performance is improved to 56.0\% (other variants of our method achieve slightly higher results). The Larsson et al. \cite{larsson2016learning} method achieves the highest performance on this metric, reaching 59.4\%. For reference, a VGG classification network fine-tuned on grayscale inputs reaches a performance of 63.5\%.

In addition to serving as a perceptual metric, this analysis demonstrates a practical use for our algorithm: without any additional training or fine-tuning, we can improve performance on grayscale image classification, simply by colorizing images with our algorithm and passing them to an off-the-shelf classifier.

\textbf{3. Raw accuracy (AuC):} As a low-level test, we compute the percentage of predicted pixel colors within a thresholded L2 distance of the ground truth in \textit{ab} color space. We then sweep across thresholds from 0 to 150 to produce a cumulative mass function, as introduced in \cite{deshpande2015learning}, integrate the area under the curve (AuC), and normalize. Note that this AuC metric measures \textit{raw prediction accuracy}, whereas our method aims for \textit{plausibility}.


Our network, trained on classification without rebalancing, outperforms our L2 variant (when trained from scratch). When the L2 net is instead fine-tuned from a color classification network, it matches the performance of the classification network. This indicates that the L2 metric can achieve accurate colorizations, but has difficulty in optimization from scratch. The Larsson et al.~\cite{larsson2016learning} method achieves slightly higher accuracy. Note that this metric is dominated by desaturated pixels, due to the distribution of \textit{ab} values in natural images (Figure \ref{fig:ab_space}(b)). As a result, even predicting gray for every pixel does quite well, and our full method with class rebalancing achieves approximately the same score.

Perceptually interesting regions of images, on the other hand, tend to have a distribution of \textit{ab} values with higher values of saturation. As such, we compute a class-balanced variant of the AuC metric by re-weighting the pixels inversely by color class probability (Equation \ref{eqn:rebal_w}, setting $\lambda=0$). Under this metric, our full method outperforms all variants and compared algorithms, indicating that class-rebalancing in the training objective achieved its desired effect.

\newfloatcommand{capbtabbox}{table}[][\FBwidth]
\begin{figure}[t]
\begin{floatrow}
\ffigbox{
\hspace{-.5 in}
 \centering
 \includegraphics[width=2.0in,height=1.82in]{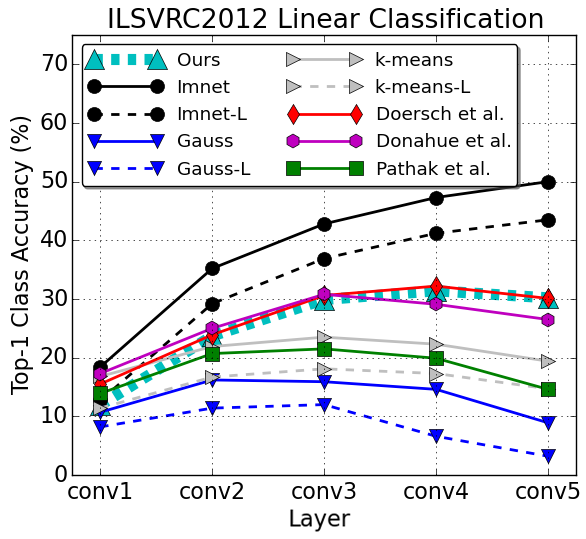}
 \label{fig:ilsvrc_lin}
}
{ 
\caption{ ImageNet Linear Classification } }

\hspace{-.5 in}
\capbtabbox{
    \centering
    \scalebox{0.7}{
        \begin{tabular}{lcccccccc}
        \specialrule{.1em}{.1em}{.1em}
        \multicolumn{9}{c}{\textbf{Dataset and Task Generalization on PASCAL \cite{everingham2010pascal}}} \\ \specialrule{.1em}{.1em}{.1em}
        \specialrule{.1em}{.1em}{.1em}
         & \multicolumn{4}{c}{\textbf{Class.}} & \multicolumn{2}{c}{\textbf{Det.}} & \multicolumn{2}{c}{\textbf{Seg.}} \\
         & \multicolumn{4}{c}{\textbf{(\%mAP)}} & \multicolumn{2}{c}{\textbf{(\%mAP)}} & \multicolumn{2}{c}{\textbf{(\%mIU)}} \\
        \specialrule{.1em}{.1em}{.1em}
        \textbf{fine-tune layers} & \textbf{[Ref]} & \textbf{fc8} & \textbf{fc6-8} & \textbf{all} & \textbf{[Ref]} & \textbf{all} & \textbf{[Ref]} & \textbf{all} \\ \hline
        ImageNet \cite{krizhevsky2012imagenet} & - & 76.8 & 78.9 & 79.9 & \cite{krahenbuhl2015data} & 56.8 & \cite{long2015fully} & 48.0 \\ \hline
        Gaussian & \cite{pathakCVPR16context} & -- & -- & 53.3 & \cite{pathakCVPR16context} & 43.4 & \cite{pathakCVPR16context} & 19.8 \\
        Autoencoder & \cite{donahue2016adversarial} & 24.8 & 16.0 & 53.8 & \cite{pathakCVPR16context} & 41.9 & \cite{pathakCVPR16context} & 25.2 \\
        k-means \cite{krahenbuhl2015data} & \cite{donahue2016adversarial} & 32.0 & 39.2 & 56.6 & \cite{krahenbuhl2015data} & 45.6 & \cite{donahue2016adversarial} & 32.6 \\ \hline
        Agrawal et al. \cite{agrawal2015learning} & \cite{donahue2016adversarial} & 31.2 & 31.0 & 54.2 & \cite{krahenbuhl2015data} & 43.9 & -- & -- \\
        Wang \& Gupta \cite{wang2015unsupervised} & -- & 28.1 & 52.2 & 58.7 & \cite{krahenbuhl2015data} & 47.4 & -- & -- \\
        *Doersch et al. \cite{doersch2015unsupervised} & \cite{donahue2016adversarial} & 44.7 & 55.1 & \textbf{65.3} & \cite{krahenbuhl2015data} & \textbf{51.1} & -- & -- \\
        *Pathak et al. \cite{pathakCVPR16context} & \cite{pathakCVPR16context} & -- & -- & 56.5 & \cite{pathakCVPR16context} & 44.5 & \cite{pathakCVPR16context} & 29.7 \\
        *Donahue et al. \cite{donahue2016adversarial} & -- & 38.2 & 50.2 & 58.6 & \cite{donahue2016adversarial} & 46.2 & \cite{donahue2016adversarial} & 34.9 \\ \hline
        Ours (gray) & -- & \textbf{52.4} & \textbf{61.5} & \textbf{65.9} & -- & 46.1 & -- & 35.0 \\
        Ours (color) & -- & \textbf{52.4} & \textbf{61.5} & \textbf{65.6} & -- & 46.9 & -- & \textbf{35.6} \\
        \specialrule{.1em}{.1em}{.1em}
        \end{tabular}
    \label{tab:pascal}
    }
}
{     \caption{ PASCAL Tests} }

\end{floatrow}


\begin{flushleft}
\textbf{Fig. 7.} \textbf{Task Generalization on ImageNet} We freeze pre-trained networks and learn linear classifiers on internal layers for ImageNet \cite{russakovsky2015imagenet} classification. Features are average-pooled, with equal kernel and stride sizes, until feature dimensionality is below 10k. ImageNet \cite{krizhevsky2012imagenet}, k-means \cite{krahenbuhl2015data}, and Gaussian initializations were run with grayscale inputs, shown with dotted lines, as well as color inputs, shown with solid lines. Previous \cite{doersch2015unsupervised,pathakCVPR16context} and concurrent \cite{donahue2016adversarial} self-supervision methods are shown. 

\textbf{Tab. 2.} \textbf{Task and Dataset Generalization on PASCAL} Classification and detection on PASCAL VOC 2007 \cite{pascal-voc-2007} and segmentation on PASCAL VOC 2012 \cite{pascal-voc-2012}, using standard mean average precision (mAP) and mean intersection over union (mIU) metrics for each task. We fine-tune our network with grayscale inputs (gray) and color inputs (color). Methods noted with a * only pre-trained a subset of the AlexNet layers. The remaining layers were initialized with \cite{krahenbuhl2015data}. Column \textbf{Ref} indicates the source for a value obtained from a previous paper.

\vspace{-.2in}

\end{flushleft}

\end{figure}

\subsection{Cross-Channel Encoding as Self-Supervised Feature Learning}
\label{sec:exp-feat}

In addition to making progress on the graphics task of colorization, we evaluate how colorization can serve as a pretext task for representation learning. Our model is akin to an autoencoder, except that the input and output are different image channels, suggesting the term {\em cross-channel encoder}. 

To evaluate the feature representation learned through this kind of cross-channel encoding, we run two sets of tests on our network. First, we test the \textit{task generalization} capability of the features by fixing the learned representation and training linear classifiers to perform object classification on already seen data (Figure 7). Second, we fine-tune the network on the PASCAL dataset \cite{everingham2010pascal} for the tasks of classification, detection, and segmentation. Here, in addition to testing on held-out \textit{tasks}, this group of experiments tests the learned representation on \textit{dataset generalization}. To fairly compare to previous feature learning algorithms, we retrain an AlexNet~\cite{krizhevsky2012imagenet} network on the colorization task, using our full method, for 450k iterations. We find that the resulting learned representation achieves higher performance on object classification and segmentation tasks relative to previous methods tested (Table 2).

\textbf{ImageNet classification} The network was pre-trained to colorize images from the ImageNet dataset, without semantic label information. We test how well the learned features represent the object-level semantics. To do this, we freeze the weights of the network, provide semantic labels, and train linear classifiers on each convolutional layer. The results are shown in Figure 7.

AlexNet directly trained on ImageNet classification achieves the highest performance, and serves as the ceiling for this test. Random initialization, with Gaussian weights or the k-means scheme implemented in \cite{krahenbuhl2015data}, peak in the middle layers. Because our representation is learned on grayscale images, the network is handicapped at the input. To quantify the effect of this loss of information, we fine-tune AlexNet on grayscale image classification, and also run the random initialization schemes on grayscale images. Interestingly, for all three methods, there is a 6\% performance gap between color and grayscale inputs, which remains approximately constant throughout the network.

We compare our model to other recent self-supervised methods pre-trained on ImageNet~\cite{doersch2015unsupervised,pathakCVPR16context,donahue2016adversarial}. To begin, our \texttt{conv1} representation results in worse linear classification performance than competiting methods~\cite{doersch2015unsupervised,donahue2016adversarial}, but is comparable to other methods which have a grayscale input. However, this performance gap is immediately bridged at \texttt{conv2}, and our network achieves competitive performance to \cite{doersch2015unsupervised,donahue2016adversarial} throughout the remainder of the network. This indicates that despite the input handicap, solving the colorization task encourages representations that linearly separate semantic classes in the trained data distribution.

\textbf{PASCAL classification, detection, and segmentation} We test our model on the commonly used self-supervision benchmarks on PASCAL classification, detection, and segmentation, introduced in~\cite{doersch2015unsupervised,krahenbuhl2015data,pathakCVPR16context}. Results are shown in Table 2. Our network achieves strong performance across all three tasks, and state-of-the-art numbers in classification and segmentation. We use the method from \cite{krahenbuhl2015data}, which rescales the layers so they ``learn" at the same rate. We test our model in two modes: (1) keeping the input grayscale by disregarding color information (Ours (gray)) and (2) modifying \texttt{conv1} to receive a full 3-channel $Lab$ input, initializing the weights on the $ab$ channels to be zero (Ours (color)).

We first test the network on PASCAL VOC 2007~\cite{pascal-voc-2007} classification, following the protocol in \cite{donahue2016adversarial}. The network is trained by freezing the representation up to certain points, and fine-tuning the remainder.
Note that when \texttt{conv1} is frozen, the network is effectively only able to interpret grayscale images. Across all three classification tests, we achieve state-of-the-art accuracy.

We also test detection on PASCAL VOC 2007, using Fast R-CNN \cite{girshick2015fast}, following the procedure in \cite{krahenbuhl2015data}. Doersch et al. \cite{doersch2015unsupervised} achieves 51.1\%, while we reach 46.9\% and 47.9\% with grayscale and color inputs, respectively. Our method is well above the strong k-means \cite{krahenbuhl2015data} baseline of 45.6\%, but all self-supervised methods still fall short of pre-training with ImageNet semantic supervision, which reaches 56.8\%.

Finally, we test semantic segmentation on PASCAL VOC 2012~\cite{pascal-voc-2012}, using the FCN architecture of~\cite{long2015fully}, following the protocol in~\cite{pathakCVPR16context}. Our colorization task shares similarities to the semantic segmentation task, as both are per-pixel classification problems. Our grayscale fine-tuned network achieves performance of 35.0\%, approximately equal to Donahue et al.~\cite{donahue2016adversarial}, and adding in color information increases performance to 35.6\%, above other tested algorithms.

\subsection{Legacy Black and White Photos}
\label{sec:comp_heldout}

\begin{figure}[t]
\centering
\includegraphics[width=1\hsize]{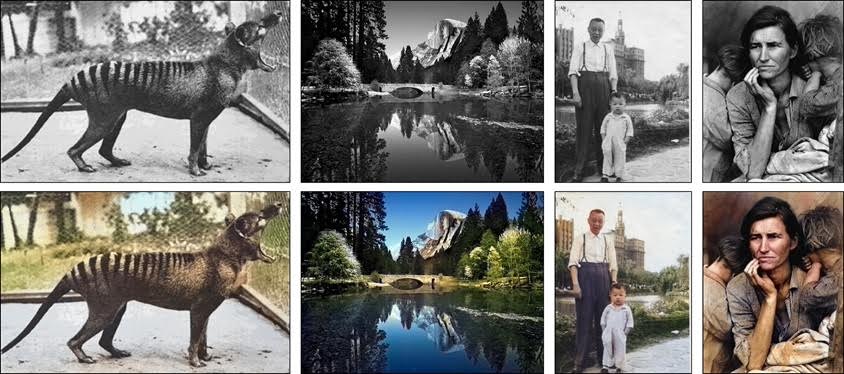}
\caption{Applying our method to legacy black and white photos. Left to right: photo by David Fleay of a Thylacine, now extinct, 1936; photo by Ansel Adams of Yosemite; amateur family photo from 1956; \textit{Migrant Mother} by Dorothea Lange, 1936.}
\label{fig:legacy}
\end{figure}

Since our model was trained using ``fake'' grayscale images generated by stripping $ab$ channels from color photos, we also ran our method on real legacy black and white photographs, as shown in Figure~\ref{fig:legacy} (additional results can be viewed on our project webpage).
One can see that our model is still able to produce good colorizations, even though the low-level image statistics of the legacy photographs are quite different from those of the modern-day photos on which it was trained.

\section{Conclusion}
While image colorization is a boutique computer graphics task, it is also an instance of a difficult pixel prediction problem in computer vision. 
Here we have shown that colorization with a deep CNN and a well-chosen objective function can come closer to producing results indistinguishable from real color photos. Our method not only provides a useful graphics output, but can also be viewed as a pretext task for representation learning. Although only trained to color, our network learns a representation that is surprisingly useful for object classification, detection, and segmentation, performing strongly compared to other self-supervised pre-training methods.

\footnotesize{\section*{Acknowledgements}
This research was supported, in part, by ONR MURI N000141010934, NSF SMA-1514512, an Intel research grant, and a hardware donation by NVIDIA Corp. We thank members of the Berkeley Vision Lab and Aditya Deshpande for helpful discussions, Philipp Kr\"ahenb\"uhl and Jeff Donahue for help with self-supervision experiments, and Gustav Larsson for providing images for comparison to \cite{larsson2016learning}.}

\bibliographystyle{splncs}
\bibliography{egbib.bib}

\section*{Appendix}

The main paper is our ECCV 2016 camera ready submission. All networks were re-trained from scratch, and are referred to as the \textbf{v2} model. Due to space constraints, we were unable to include many of the analyses presented in our original \textbf{arXiv v1} paper. We include these analyses in this Appendix, which were generated from a previous \textbf{v1} version of the model. All models are publicly available on our website.

Section \ref{sec:exp-feat-supp} contains additional representation learning experiments. Section \ref{sec:vgg_eval} investigates additional analysis on the VGG semantic interpretability test. In Section \ref{sec:res_low}, we explore how low-level queues affect the output. Section \ref{sec:res_multi} examines the multi-modality learned in the network.
Section \ref{sec:net-arch} defines the network architecture used. In Section \ref{sec:held-out}, we compare our algorithm to previous approaches \cite{deshpande2015learning} and \cite{cheng2015deep}, and show additional examples on legacy grayscale images.

\begin{table}[t!]
\centering
\scalebox{0.75} {
\begin{tabular}{c@{\hskip.07in}c@{\hskip.07in}c@{\hskip.15in}c@{\hskip.07in}c@{\hskip.07in}c@{\hskip.07in}@{\hskip.07in}c@{\hskip.07in}c@{\hskip.07in}c@{\hskip.07in}c@{\hskip.07in}c}
\specialrule{.1em}{.1em}{.1em}
\multirow{2}{*}{\textbf{Author}} & \multirow{2}{*}{\textbf{Training}} & \multirow{2}{*}{\textbf{Input}} & \multicolumn{3}{c}{\textbf{Model}} & \multirow{2}{*}{\textbf{[Ref]}} & \multicolumn{4}{c}{\textbf{Layers}} \\
 &  &  & \textbf{Params} & \textbf{Feats} & \textbf{Runtime} & & \textbf{conv2} & \textbf{conv3} & \textbf{conv4} & \textbf{conv5} \\
\specialrule{.1em}{.1em}{.1em}
\specialrule{.1em}{.1em}{.1em}
Krizhevsky et al. \cite{krizhevsky2012imagenet} & labels & rgb & 1.00 & 1.00 & 1.00 & -- & 56.5 & 56.5 & 56.5 & 56.5 \\
Krizhevsky et al. \cite{krizhevsky2012imagenet} & labels & L & 0.99 & 1.00 & 0.92 & -- & 50.5 & 50.5 & 50.5 & 50.5 \\ \hline
Noroozi \& Favaro \cite{noroozi2016unsupervised} & imagenet & rgb & 1.00 & 5.60 & 7.35 & \cite{noroozi2016unsupervised} & 56.0 & 52.4 & 48.3 & 38.1 \\ \hline
Gaussian & imagenet & rgb & 1.00 & 1.00 & 1.00 & \cite{noroozi2016unsupervised} & 41.0 & 34.8 & 27.1 & 12.0 \\ \hline
Doersch et al. \cite{doersch2015unsupervised} & imagenet & rgb & 1.61 & 1.00 & 2.82 & \cite{noroozi2016unsupervised} & 47.6 & \textbf{48.7} & \textbf{45.6} & 30.4 \\
Wang \& Gupta \cite{wang2015unsupervised} & videos & rgb & 1.00 & 1.00 & 1.00 & \cite{noroozi2016unsupervised} & 46.9 & 42.8 & 38.8 & 29.8 \\
Donahue et al. \cite{donahue2016adversarial} & imagenet & rgb & 1.00 & 0.87 & 0.96 & \cite{donahue2016adversarial} & \textbf{51.9} & 47.3 & 41.9 & 31.1 \\ \hline
Ours & imagenet & L & 0.99 & 0.87 & 0.84 & -- & 46.6 & 43.5 & 40.7 & \textbf{35.2} \\
\specialrule{.1em}{.1em}{.1em}
\end{tabular} }
\caption{\textbf{ImageNet classification with nonlinear layers}, as proposed in \cite{noroozi2016unsupervised}. Note that some models have architectural differences. We note the effect of these modifications by the number of model parameters, number of features per image, and run-time, as a multiple of Alexnet \cite{krizhevsky2012imagenet} without modifications, up to the \texttt{pool5} layer. Noroozi et al. \cite{noroozi2016unsupervised} performs best on all layers, with denser feature maps due to smaller stride in \texttt{conv1} layer, along with \texttt{LRN} and \texttt{pool} ordering switched. Doersch et al. \cite{doersch2015unsupervised} remove groups in \texttt{conv} layers. Donahue et al. \cite{donahue2016adversarial} remove \texttt{LRN} layers and change \texttt{ReLU} to \texttt{leakyReLU} units. Ours removes \texttt{LRN} and uses a single channel input. We also note the source of performance numbers. Column \textbf{Ref} indicates the source for a value obtained from a previous paper.}
\label{tab:ilsvrc-nonlin}
\end{table}

\section{Cross-Channel Encoding as Self-Supervised Feature Learning (continued)}
\label{sec:exp-feat-supp}

In Section \ref{sec:exp-feat}, we discussed using colorization as a pretext task for representation learning. In addition to learning \textit{linear} classifiers on internal layers for ImageNet classifiers, we run the additional experiment of learning \textit{non-linear} classifiers, as proposed in \cite{noroozi2016unsupervised}. Each internal layer is frozen, along with all preceding layers, and the layers on top are randomly reinitialized and trained for classification. Performance is summarized in Table \ref{tab:ilsvrc-nonlin}. Of the unsupervised models, Noroozi et al. \cite{noroozi2016unsupervised} have the highest performance across all layers. The architectural modifications result in $5.6\times$ feature map size and $7.35\times$ model run-time, up to the \texttt{pool5} layer, relative to an unmodified Alexnet. Of the remaining methods, Donahue et al. \cite{donahue2016adversarial} performs best at \texttt{conv2} and Doersch et al. performs best at \texttt{conv3} and \texttt{conv4}. Our method performs strongly throughout, and best across methods at the \texttt{conv5} layer.

\section{Semantic Interpretability of Colorizations}
\label{sec:vgg_eval}

In Section \ref{sec:exp-imnet}, we investigated using the VGG classifier to evaluate the semantic interpretability of our colorization results. In Section \ref{sec:vgg_topcats}, we show the categories which perform well, and the ones which perform poorly, using this metric. In Section \ref{sec:vgg_conf}, we show commonly confused categories after recolorization.

\subsection{Category Performance}
\label{sec:vgg_topcats}

\begin{figure}[t!]
 \centering
 \includegraphics[width=1.\hsize]{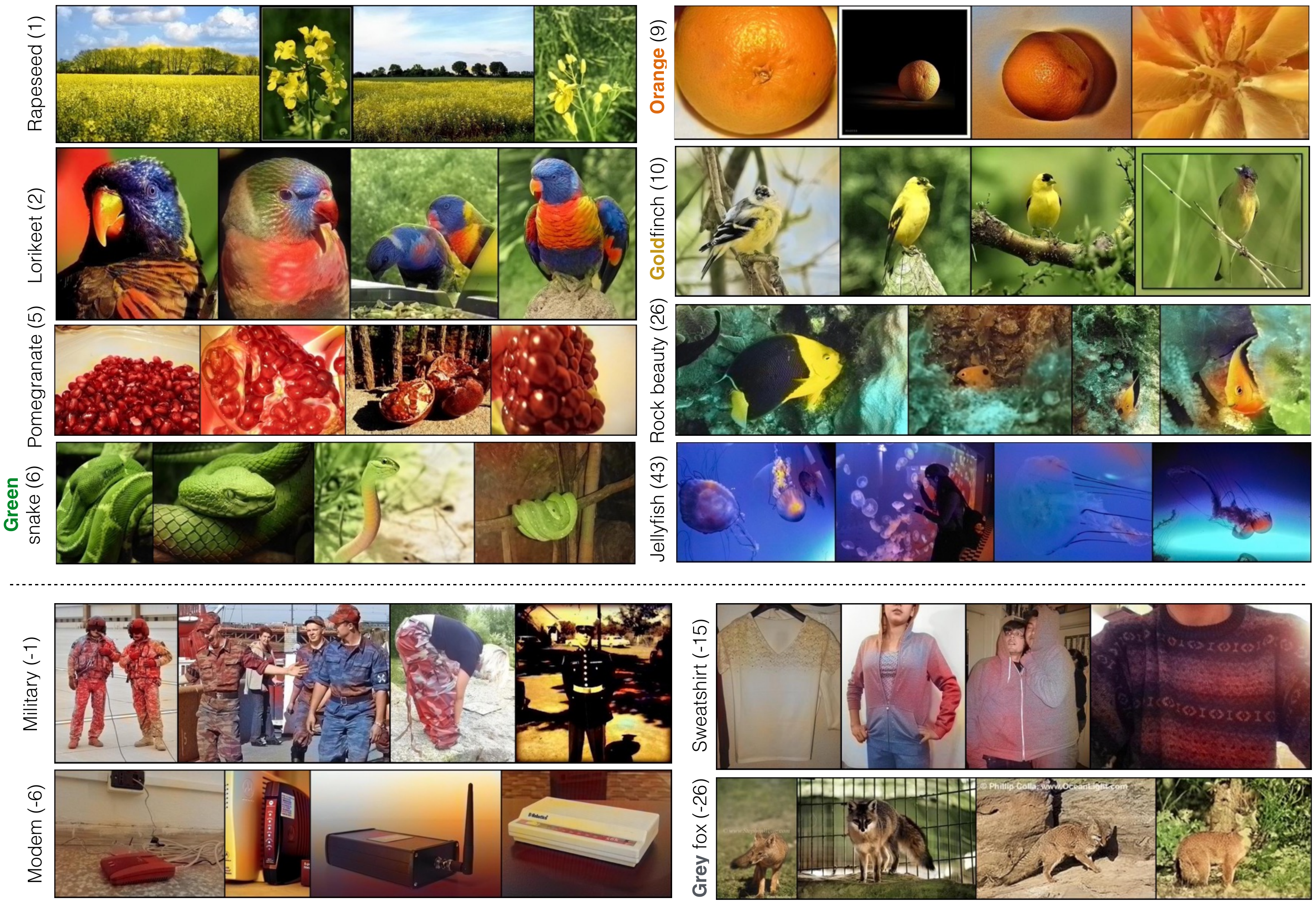}
 \vspace{-.25in}
 \caption{Images colorized by our algorithm from selected categories. Categories are sorted by VGG object classification accuracy of our colorized images relative to accuracy on gracyscale images. Top: example categories where our colorization \textit{helps} the most. Bottom: example categories where our colorization \textit{hurts} the most. Number in parentheses indicates category rank amongst all 1000. Notice that the categories most affected by colorization are those for which color information is highly diagnostic, such as birds and fruits. The bottom examples show several kinds of failures: 1) artificial objects such as modems and clothes have ambiguous colors; color is not very informative for classification, and moreover, our algorithm tends to predict an incoherent distribution of red and blue, 2) for certain categories, like the gray fox, our algorithm systematically predicts the wrong color, confusing the species.}
 \label{fig:vgg_res_examples}
\end{figure}

In Figure~\ref{fig:vgg_res_examples}, we show a selection of classes that have the most improvement in VGG classification with respect to grayscale, along with the classes for which our colorizations hurt the most. Interestingly, many of the top classes actually have a color in their name, such as the green snake, orange, and goldfinch. The bottom classes show some common errors of our system, such as coloring clothing incorrectly and inconsistently and coloring an animal with a plausible but incorrect color. This analysis was performed using 48k images from the ImageNet validation set, and images in the top and bottom 10 classes are provided on the website.

Our process for sorting categories and images is described below. For each category, we compute the top-5 classification performance on grayscale and recolorized images, $\mathbf{a}_{gray}, \mathbf{a}_{recolor} \in [0,1]^{C}$, where $C=1000$ categories. We sort the categories by $\mathbf{a}_{recolor}-\mathbf{a}_{gray}$. The re-colored vs grayscale performance per category is shown in Figure \ref{fig:vgg_plt}(a), with top and bottom 50 categories highlighted. For the top example categories, the individual images are sorted by ascending rank of the correct classification of the recolorizeed image, with tiebreakers on descending rank of the correct classification of the grayscale image. For the bottom example categories, the images are sorted in reverse, in order to highlight the instances when recolorization results in an errant classification relative to the grayscale image.

\subsection{Common Confusions}
\label{sec:vgg_conf}

\begin{figure}[H]
 \centering
 \includegraphics[width=1.\hsize]{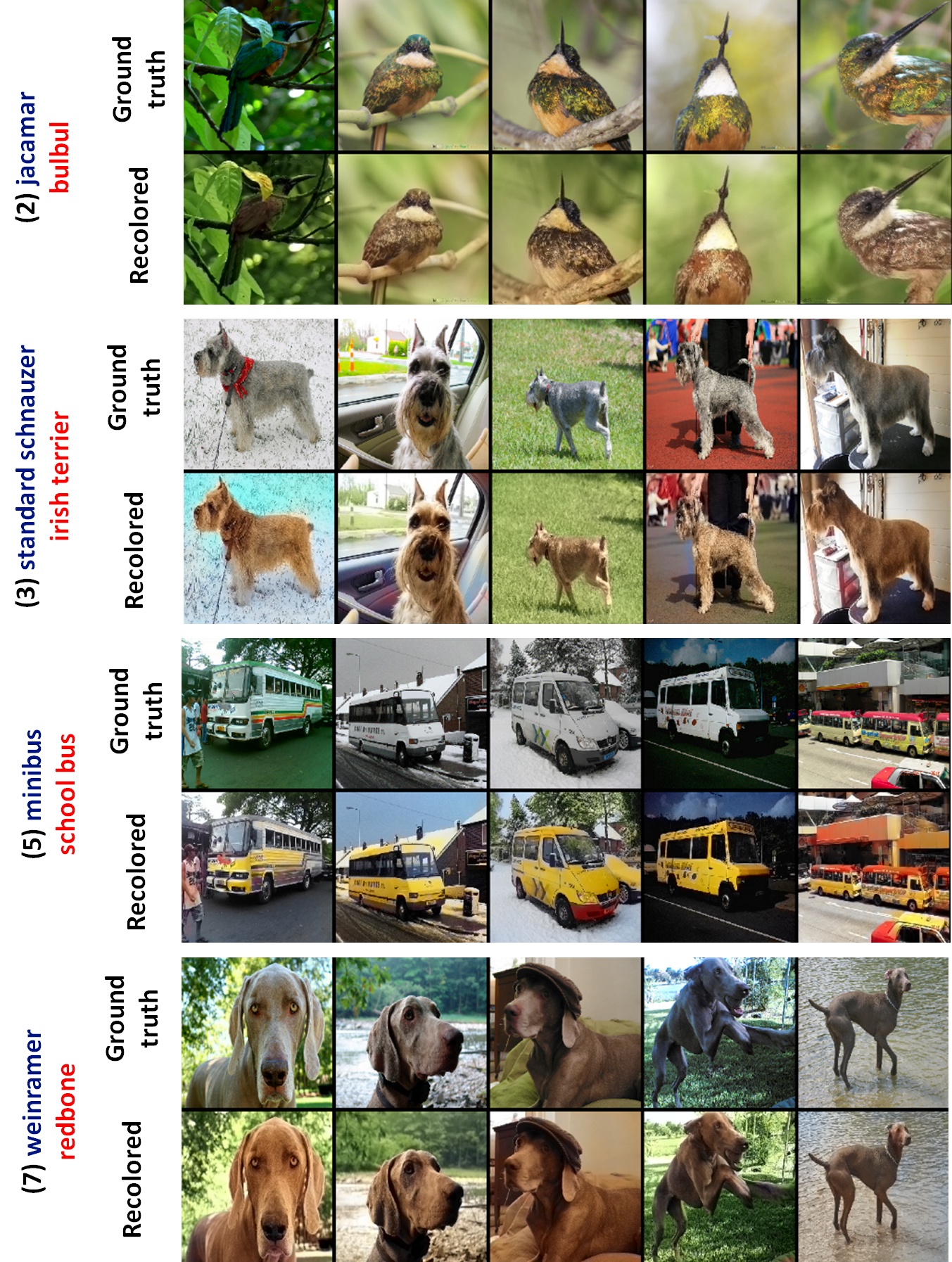}
  \caption{Examples of some most-confused categories. Top rows show ground truth image. Bottom rows show recolorized images. Rank of common confusion in parentheses. \textcolor{MyDarkBlue}{\textbf{Ground truth}} and \textcolor{red}{\textbf{confused}} categories after recolorization are labeled.}
 \label{fig:comm_conf_mont}
\end{figure}

To further investigate the biases in our system, we look at the common classification confusions that often occur after image recolorization, but not with the original ground truth image. Examples for some top confusions are shown in Figure \ref{fig:comm_conf_mont}. An image of a ``minibus" is often colored yellow, leading to a misclassification as ``school bus". Animal classes are sometimes colored differently than ground truth, leading to misclassification to related species. Note that the colorizations are often visually realistic, even though they lead to a misclassification.

\begin{figure}[t!]
 \centering
 \includegraphics[width=1.0\hsize]{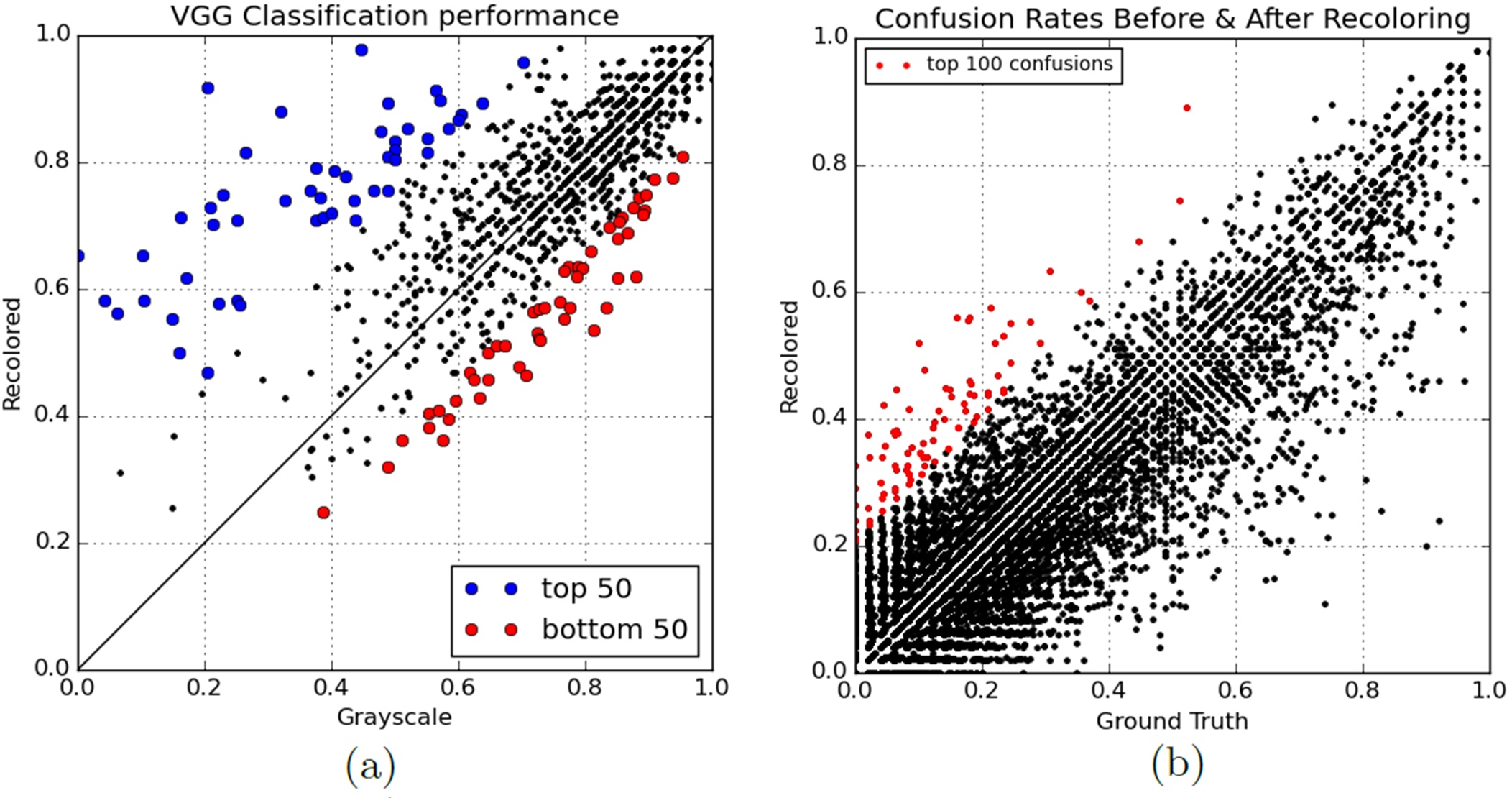}
  \caption{(a) Performance of VGG top-5 classification on recolorized images vs grayscale images per category (b) Top-5 confusion rates with recolorizations and original colors. Test was done on last 48,000 images in ImageNet validation set.}
 \label{fig:vgg_plt}
\end{figure}

To find common confusions, we compute the rate of top-5 confusion $\mathbf{C}_{orig},\mathbf{C}_{recolor}\in[0,1]^{C\times C}$, with ground truth colors and after recolorization. A value of $\mathbf{C}_{c,d}=1$ means that every image in category $c$ was classified as category $d$ in the top-5.
We find the class-confusion added after recolorization by computing $\mathbf{A}=\mathbf{C}_{recolor}-\mathbf{C}_{orig}$, and sort the off-diagonal entries. Figure \ref{fig:vgg_plt}(b) shows all $C\times (C-1)$ off-diagonal entries of $\mathbf{C}_{recolor}$ vs $\mathbf{C}_{orig}$, with the top 100 entries from $\mathbf{A}$ highlighted. For each category pair $(c,d)$, we extract the images that contained the confusion after recolorization, but not with the original colorization. We then sort the images in descending order of the classification score of the confused category. 

\section{Is the network exploiting low-level cues?}
\label{sec:res_low}

Unlike many computer vision tasks that can be roughly categorized as low, mid or high-level vision, color prediction requires understanding an image at both the pixel and the semantic-level. We have investigated how colorization generalizes to high-level semantic tasks in Section \ref{sec:exp-feat}. Studies of natural image statistics have shown that the lightness value of a single pixel can highly constrain the likely color of that pixel: darker lightness values tend to be correlated with more saturated colors~\cite{chakrabarti2015color}. 

\begin{figure}[t!]
 \centering
 \includegraphics[width=1.0\hsize]{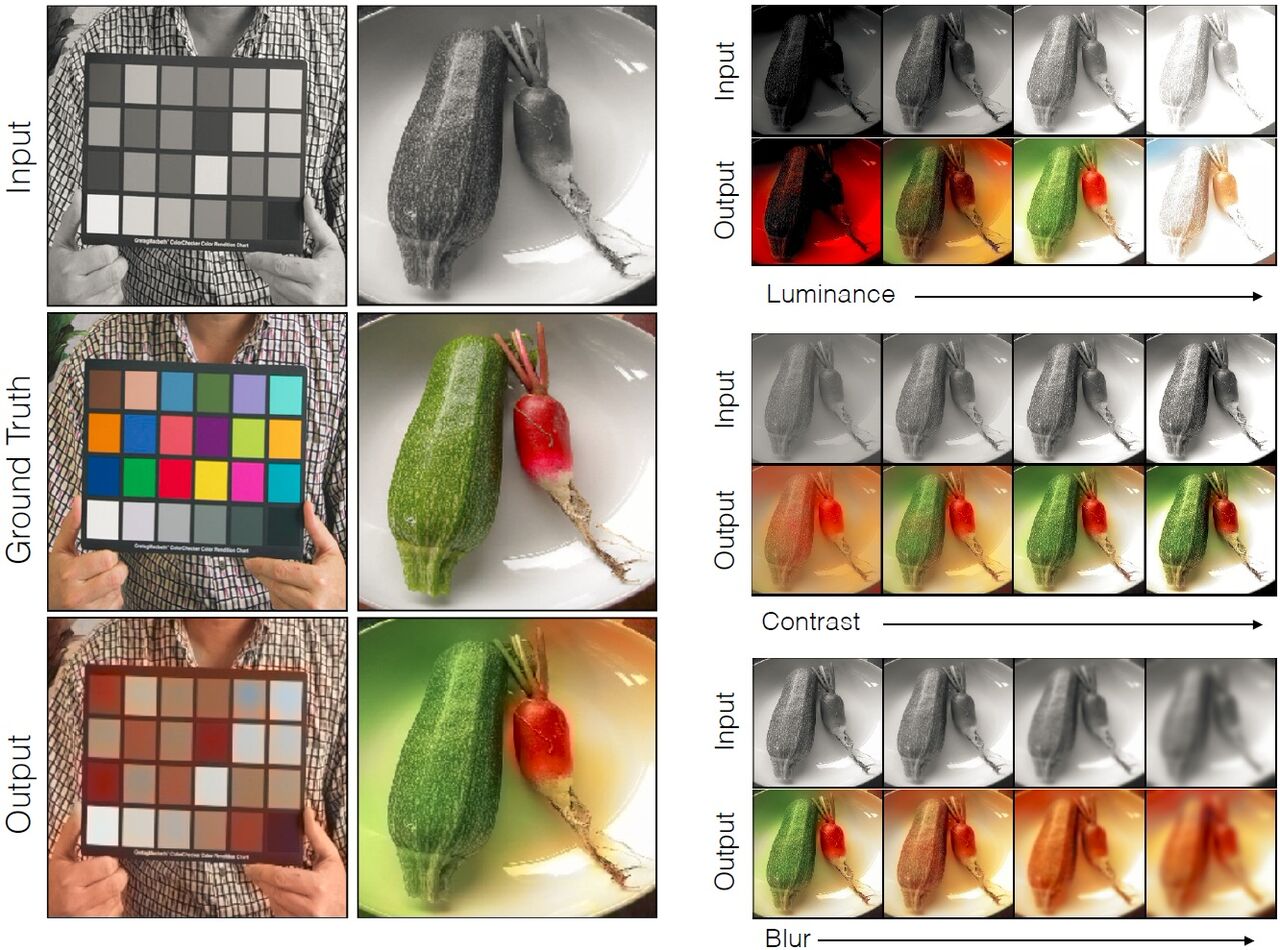}
 \caption{Left: pixel lightness on its own does not reveal color, as shown by the color chart. In contrast, two vegetables that are nearly isoluminant are recognized as having different colors. Right: stability of the network predictions with respect to low-level image transformations.}
 \label{fig:low_level_examples}
\end{figure}

Could our network be exploiting a simple, low-level relationship like this, in order to predict color?\footnote{E.g., previous work showed that CNNs can learn to use chromatic aberration cues to predict, given an image patch, its (x,y) location within an image~\cite{doersch2015unsupervised}.}
We tested this hypothesis with the simple demonstration in Figure~\ref{fig:low_level_examples}. Given a grayscale Macbeth color chart as input, our network was unable to recover its colors. This is true, despite the fact that the lightness values vary considerably for the different color patches in this image. On the other hand, given two recognizable vegetables that are roughly isoluminant, the system is able to recover their color.

In Figure \ref{fig:low_level_examples}, we also demonstrate that the prediction is somewhat stable with respect to low-level lightness and contrast changes. Blurring, on the other hand, has a bigger effect on the predictions in this example, possibly because the operation removes the diagnostic texture pattern of the zucchini.

\begin{figure}[t!]
 \centering
 \includegraphics[width=1.\hsize]{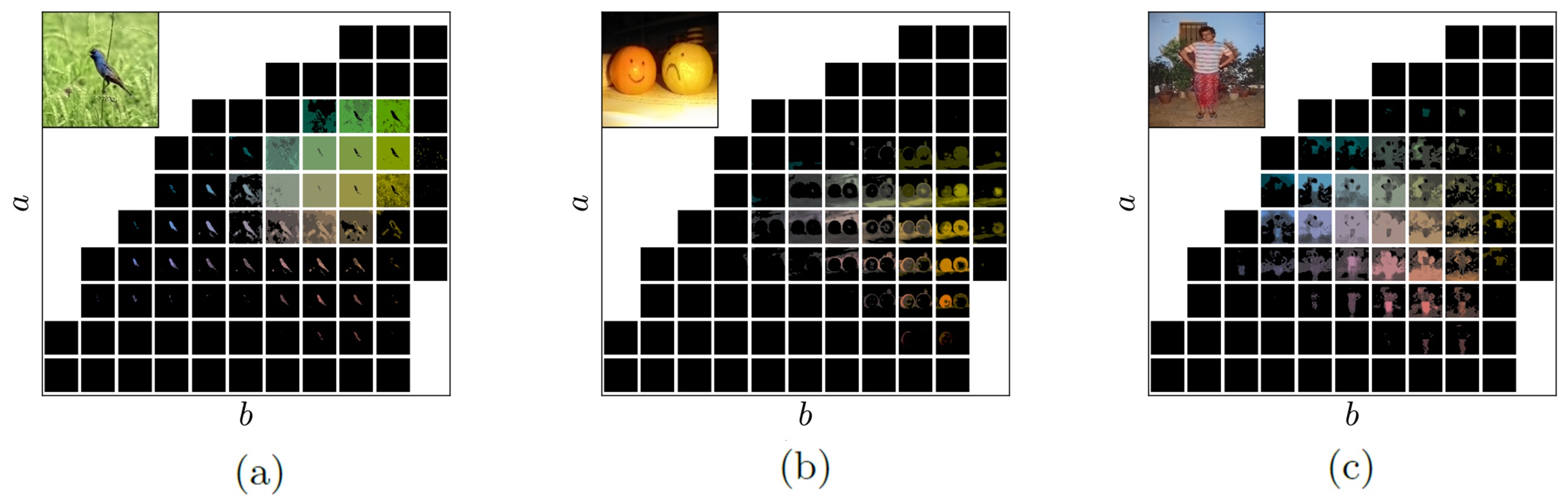}
 \caption{The output probability distributions per image. The top-left image is final prediction of our system. The black sub-images are quantized blocks of the $ab$ gamut. High probabilities are shown as higher luminance and are quantized for clarity. (a) Background of bird is predicted to be green or brown. Foreground bird has distribution across blue and red colors. (b) Oranges are predicted to be different colors. (c) The person's shirt and sarong has uncertainty across turqoise/cyan/orange and red/pink/purple colors, respectively. Note that despite the multimodality of the per-pixel distributions, the results after taking the annealed-mean are typically spatially consistent.}
 \label{fig:distr_monts}
\end{figure}

\section{Does our model learn multimodal color distributions?}
\label{sec:res_multi}

As discussed in Section \ref{sec:obj}, formulating color prediction as a multinomial classification problem allows the system to predict multimodal distributions, and can capture the inherent ambiguity in the color of natural objects. In Figure \ref{fig:distr_monts}, we illustrate the probability outputs $\mathbf{\widehat{Z}}$ and demonstrate that the network does indeed learn multimodal distributions. The system output $\mathbf{\widehat{Y}}$ is shown in the top-left of Figure \ref{fig:distr_monts}. Each block illustrates the probability map $\mathbf{\widehat{Z}}_{q}\in[0,1]^{H,W}$ given $ab$ bin $q$ in the output space. For clarity, we show a subsampling of the $Q$ total output bins and coarsely quantize the probability values. In Figure \ref{fig:distr_monts}(a), the system clearly predicts a different distribution for the background vegetation and the foreground bird. The background is predicted to be green, yellow, or brown, while the foreground bird is predicted to be red or blue. Figure \ref{fig:distr_monts}(b) shows that oranges can be predicted to be different colors. Lastly, in Figure \ref{fig:distr_monts}(c), the man's sarong is predicted to be either red, pink, or purple, while his shirt is classified as turquoise, cyan or light orange. Note that despite the multi-modality of the prediction, taking the annealed-mean of the distribution produces a spatially consistent prediction.

\section{Network architecture}
\label{sec:net-arch}
Figure \ref{fig:architecture} showed a diagram of our network architecture. Table \ref{tab:arch} in this document thoroughly lists the layers used in our architecture during training time. During testing, the temperature adjustment, softmax, mean, and bilinear upsampling are all implemented as subsequent layers in a feed-forward network. Note the column showing the effective dilation. The effective dilation is the spacing at which consecutive elements of the convolutional kernel are evaluated, relative to the input pixels, and is computed by the product of the accumulated stride and the layer dilation. Through each convolutional block from \texttt{conv1} to \texttt{conv5}, the effective dilation of the convolutional kernel is increased. From \texttt{conv6} to \texttt{conv8}, the effective dilation is decreased.

\begin{table}[t]
\centering
\scalebox{0.8} {
\begin{tabular}{@{\hskip .1in}c@{\hskip .2in}c@{\hskip .1in}c@{\hskip .1in}c@{\hskip .1in}c@{\hskip .1in}c@{\hskip .1in}c@{\hskip .1in}c@{\hskip .1in}c@{\hskip .1in}}
\specialrule{.1em}{.1em}{.1em}
 \textbf{Layer} & \textbf{X} & \textbf{C} & \textbf{S} & {\textbf{D}} & \textbf{Sa} & \textbf{De} & \textbf{BN} & \textbf{L} \\
 \specialrule{.1em}{.1em}{.1em}
 \specialrule{.1em}{.1em}{.1em}
\textbf{data} & 224 & 3 & - & - & - & - & - & - \\ \hline
 \textbf{conv1\_1} & 224 & 64 & 1 & 1 & 1 & 1 & - & - \\
\textbf{conv1\_2} & 112 & 64 & 2 & 1 & 1 & 1 & \checkmark & - \\ \hline
\textbf{conv2\_1} & 112 & 128 & 1 & 1 & 2 & 2 & - & - \\
\textbf{conv2\_1} & 56 & 128 & 2 & 1 & 2 & 2 & \checkmark & - \\ \hline
\textbf{conv3\_1} & 56 & 256 & 1 & 1 & 4 & 4 & - & - \\
\textbf{conv3\_2} & 56 & 256 & 1 & 1 & 4 & 4 & - & - \\
\textbf{conv3\_3} & 28 & 256 & 2 & 1 & 4 & 4 & \checkmark & - \\ \hline
\textbf{conv4\_1} & 28 & 512 & 1 & 1 & 8 & 8 & - & - \\
\textbf{conv4\_2} & 28 & 512 & 1 & 1 & 8 & 8 & - & - \\
\textbf{conv4\_3} & 28 & 512 & 1 & 1 & 8 & 8 & \checkmark & - \\ \hline
\textbf{conv5\_1} & 28 & 512 & 1 & 2 & 8 & 16 & - & - \\
\textbf{conv5\_2} & 28 & 512 & 1 & 2 & 8 & 16 & - & - \\
\textbf{conv5\_3} & 28 & 512 & 1 & 2 & 8 & 16 & \checkmark & - \\ \hline
\textbf{conv6\_1} & 28 & 512 & 1 & 2 & 8 & 16 & - & - \\
\textbf{conv6\_2} & 28 & 512 & 1 & 2 & 8 & 16 & - & - \\
\textbf{conv6\_3} & 28 & 512 & 1 & 2 & 8 & 16 & \checkmark & - \\ \hline
\textbf{conv7\_1} & 28 & 256 & 1 & 1 & 8 & 8 & - & - \\
\textbf{conv7\_2} & 28 & 256 & 1 & 1 & 8 & 8 & - & - \\
\textbf{conv7\_3} & 28 & 256 & 1 & 1 & 8 & 8 & \checkmark & - \\ \hline
\textbf{conv8\_1} & 56 & 128 & .5 & 1 & 4 & 4 & - & - \\
\textbf{conv8\_2} & 56 & 128 & 1 & 1 & 4 & 4 & - & - \\
\textbf{conv8\_3} & 56 & 128 & 1 & 1 & 4 & 4 & - & \checkmark \\
\specialrule{.1em}{.1em}{.1em}
\end{tabular}
}
\caption{Our network architecture. \textbf{X} spatial resolution of output, \textbf{C} number of channels of output; \textbf{S} computation stride, values greater than 1 indicate downsampling following convolution, values less than 1 indicate upsampling preceding convolution; \textbf{D} kernel dilation; \textbf{Sa} accumulated stride across all preceding layers (product over all strides in previous layers); \textbf{De} effective dilation of the layer with respect to the input (layer dilation times accumulated stride); \textbf{BN} whether \texttt{BatchNorm} layer was used after layer; \textbf{L} whether a 1x1 \texttt{conv} and cross-entropy loss layer was imposed}
\label{tab:arch}
\end{table}

\section{Colorization comparisons on held-out datasets}
\label{sec:held-out}

\subsection{Comparison to LEARCH \cite{deshpande2015learning}}
\label{sec:comp_heldout}

\begin{figure}[t]
\includegraphics[width=.6\linewidth]{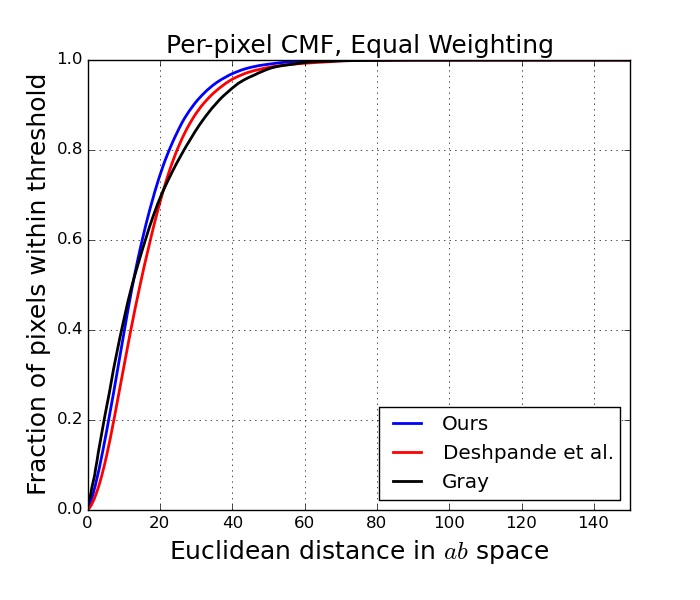}
\caption{CMF on the LEARCH~\cite{deshpande2015learning} test set}
\label{fig:res_learch_cmf}
\end{figure}

Though our model was trained on object-centric ImageNet dataset, we demonstrate that it nonetheless remains effective for photos from the scene-centric SUN dataset~\cite{xiao2010sun} selected by Deshpande et al.~\cite{deshpande2015learning}. Deshpande et al. recently established a benchmark for colorization using a subset of the SUN dataset and reported top results using an algorithm based on LEARCH~\cite{ratliff2009learning}. Table \ref{table:res_learch} provides a quantitative comparison of our method to Deshpande et al.. For fair comparison, we use the same grayscale input as \cite{deshpande2015learning}, which is $\frac{R+G+B}{3}$. Note that this input space is non-linearly related to the $L$ channel on which we trained. Despite differences in grayscale space and training dataset, our method outperforms Deshpande et al. in both the raw accuracy AuC CMF and perceptual realism AMT metrics. Figure \ref{fig:res_learch_cmf} shows qualitative comparisons between our method and Deshpande et al., one from each of the six scene categories. A complete comparison on all 240 images are included in the supplementary material. Our results are able to fool participants in the \textit{real vs. fake} task $17.2\%$ of the time, significantly higher than Deshpande et al. at $9.8\%$.

\begin{table}[t]
\begin{tabular}{l@{\hskip .05in}c@{\hskip .05in}c}
\multicolumn{3}{c}{Results on LEARCH \cite{deshpande2015learning} dataset} \\
\specialrule{.1em}{.1em}{.1em}
    & AuC & AMT \\
Algorithm & CMF & Labeled \\
    & (\%) & Real (\%) \\  \hline
Ours & {\bf 90.1} & {\bf 17.2$\pm$1.9}  \\
Deshpande et al. \cite{deshpande2015learning} & 88.8 & 9.8$\pm$1.5 \\
Grayscale & 89.3 & --  \\  \hline
Ground Truth & 100 & 50  \\
\specialrule{.1em}{.1em}{.1em}
\end{tabular}
\caption{Results on LEARCH~\cite{deshpande2015learning} test set, containing 240 images from 6 categories \textit{beach, outdoor, castle, bedroom, kitchen,} and \textit{living room}. Results column 1 shows the AuC of thresholded CMF over ~\textit{ab} space. Results column 2 are from our AMT real vs. fake test.}
\label{table:res_learch}
\end{table}

\begin{figure}[t!]
 \centering
 \includegraphics[width=1.\hsize]{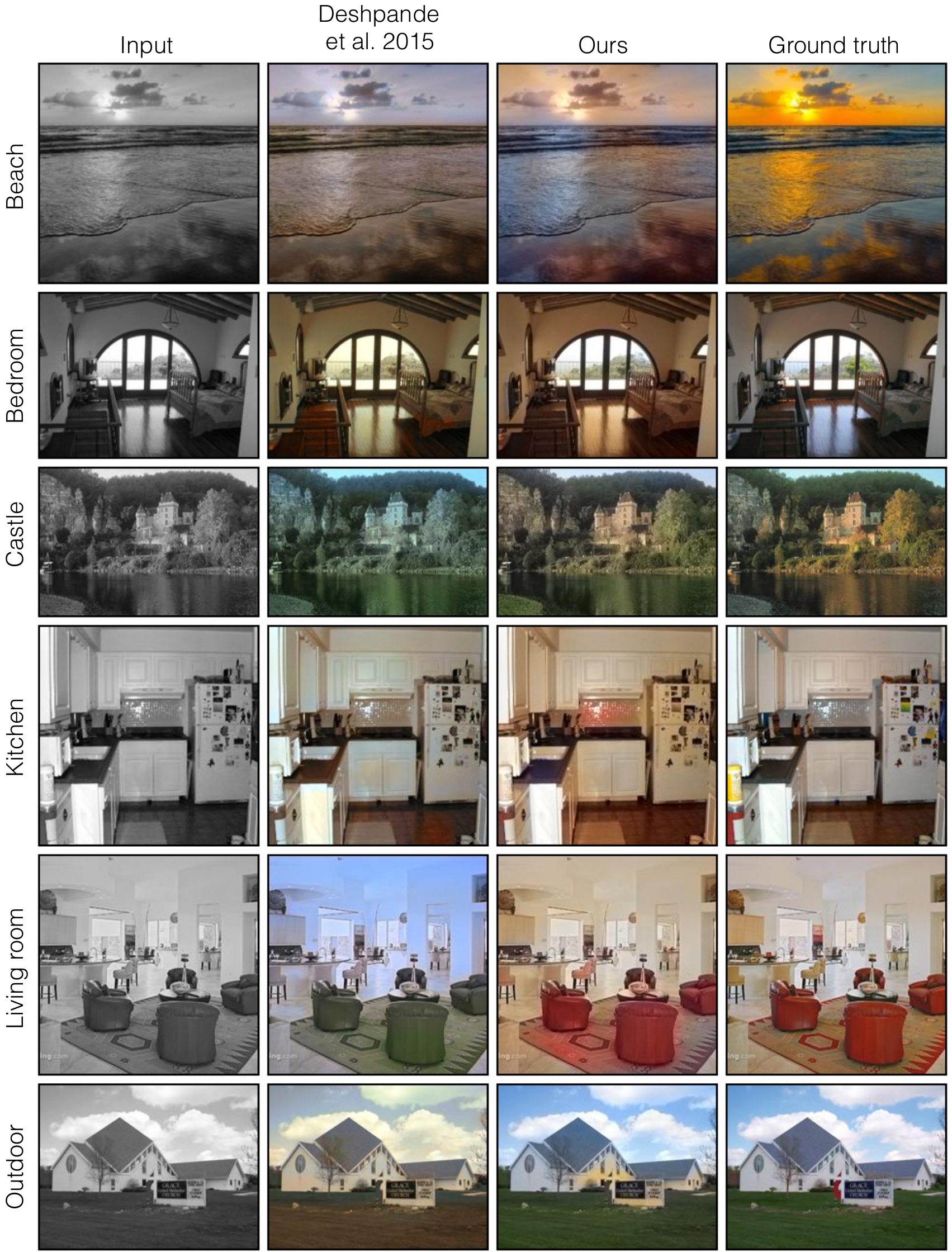}
 \vspace{-0.1in}
  \caption{Our model generalizes well to datasets on which it was not trained. Here we show results on the dataset from \cite{deshpande2015learning}, which consists of six scene categories from SUN \cite{xiao2010sun}. Compared to the state of the art algorithm on this dataset \cite{deshpande2015learning}, our method produces more perceptually plausible colorization (see also Table \ref{table:res_learch} and Figure \ref{fig:res_learch_cmf}). Please visit \texttt{http://richzhang.github.io/colorization/} to see the results on all 240 images.}
 \label{fig:learch_comparison}
  \vspace{-0.1in}
\end{figure}

\subsection{Comparison to Deep Colorization \cite{cheng2015deep}}
\label{sec:add_ex}

We provide qualitative comparisons to the 23 test images in \cite{cheng2015deep} on the website, which we obtained by manually cropping from the paper. Our results are about the same qualitative level as \cite{cheng2015deep}. Note that Deep Colorization \cite{cheng2015deep} has several advantages in this setting: (1) the test images are from the SUN dataset \cite{patterson2012sun}, which we did not train on and (2) the 23 images were hand-selected from 1344 by the authors, and is not necessarily representative of algorithm performance. We were unable to obtain the 1344 test set results through correspondence with the authors.

Additionally, we compare the methods on several important dimensions in Table \ref{tab:comp_dc}: algorithm pipeline, learning, dataset, and run-time. Our method is faster, straightforward to train and understand, has fewer hand-tuned parameters and components, and has been demonstrated on a broader and more diverse set of test images than Deep Colorization \cite{cheng2015deep}.

\begin{table}[t!]
\centering
\scalebox{1.} {
\begin{tabular}{c@{\hskip .2in}c@{\hskip .2in}c}
\specialrule{.1em}{.1em}{.1em}
 & {\textbf{Deep Colorization} \cite{cheng2015deep}} & {\textbf{Ours}} \\
\specialrule{.1em}{.1em}{.1em}
\specialrule{.1em}{.1em}{.1em}
 & {(1) Extract feature sets} & {} \\
 & {(a) 7x7 patch (b) DAISY} & {} \\
{\textbf{Algorithm}} & {(c) FCN on 47 categories} & {Feed-forward CNN} \\
 & {(2) 3-layer NN regressor} & {} \\
 & {(3) Joint-bilateral filter} & {} \\ \hline
 & {Extract features. Train FCN \cite{long2015fully}} & {Train CNN from pixels to} \\
{\textbf{Learning}} & {on pre-defined categories.} & {color distribution. Tune single} \\
 & {Train 3-layer NN regressor.} & {parameter on validation.} \\ \hline
 & {2688/1344 images from} & {1.3M/10k images from} \\
{\textbf{Dataset}} & {SUN \cite{patterson2012sun} for train/test.} & {ImageNet \cite{russakovsky2015imagenet} for train/test.} \\
 & {Limited variety with} & {Broad and diverse} \\
 & {only scenes.} & {set of objects and scenes.} \\ \hline
 & {4.9s/image on} & {21.1ms/image in \textit{Caffe}} \\
{\textbf{Run-time}} & {Matlab implementation} & {on K40 GPU} \\
\specialrule{.1em}{.1em}{.1em}
\end{tabular}
}
\caption{Comparison to Deep Colorization \cite{cheng2015deep}}
\label{tab:comp_dc}
\end{table}

\subsection{Additional Examples on Legacy Grayscale Images}

\begin{figure}
 \centering
 \includegraphics[width=.95\hsize]{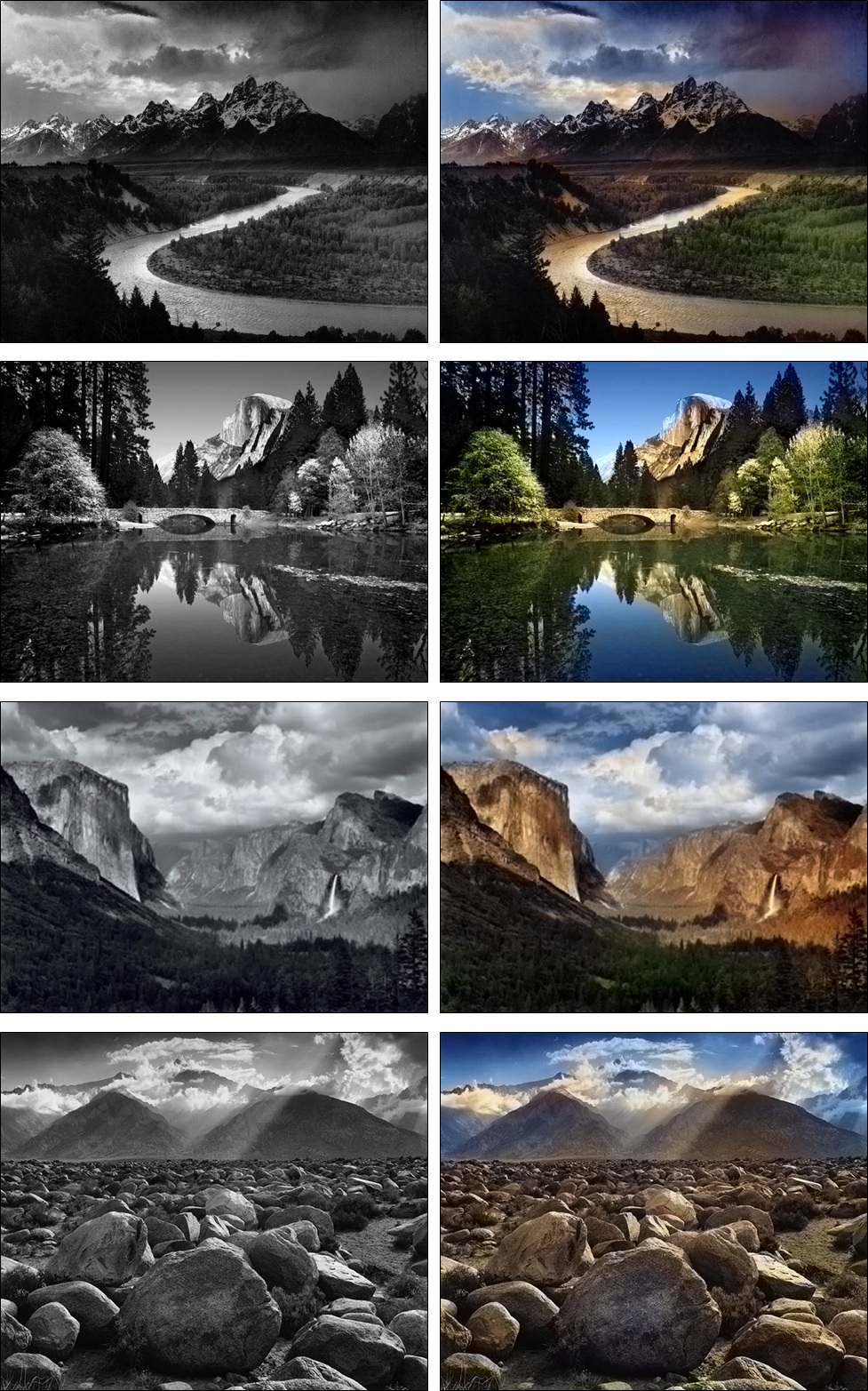}
  \caption{Applying our method to black and white photographs by Ansel Adams.}
 \label{fig:famous_examples_adams}
\end{figure}

\begin{figure}
 \centering
 \includegraphics[width=1\hsize]{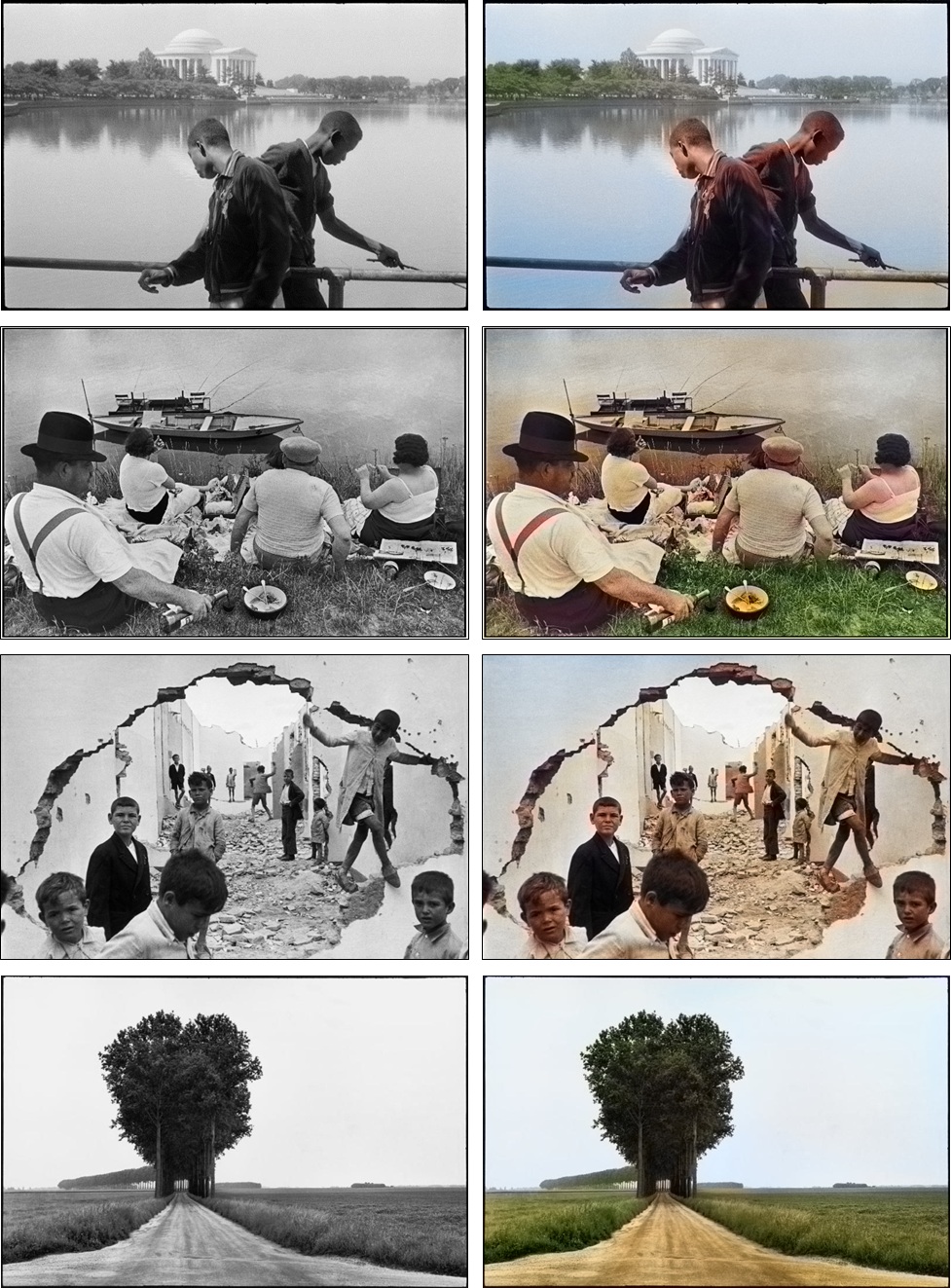}
  \caption{Applying our method to black and white photographs by Henri Cartier-Bresson.}
 \label{fig:famous_examples_bresson}
\end{figure}

\begin{figure}
 \centering
 \includegraphics[width=1\hsize]{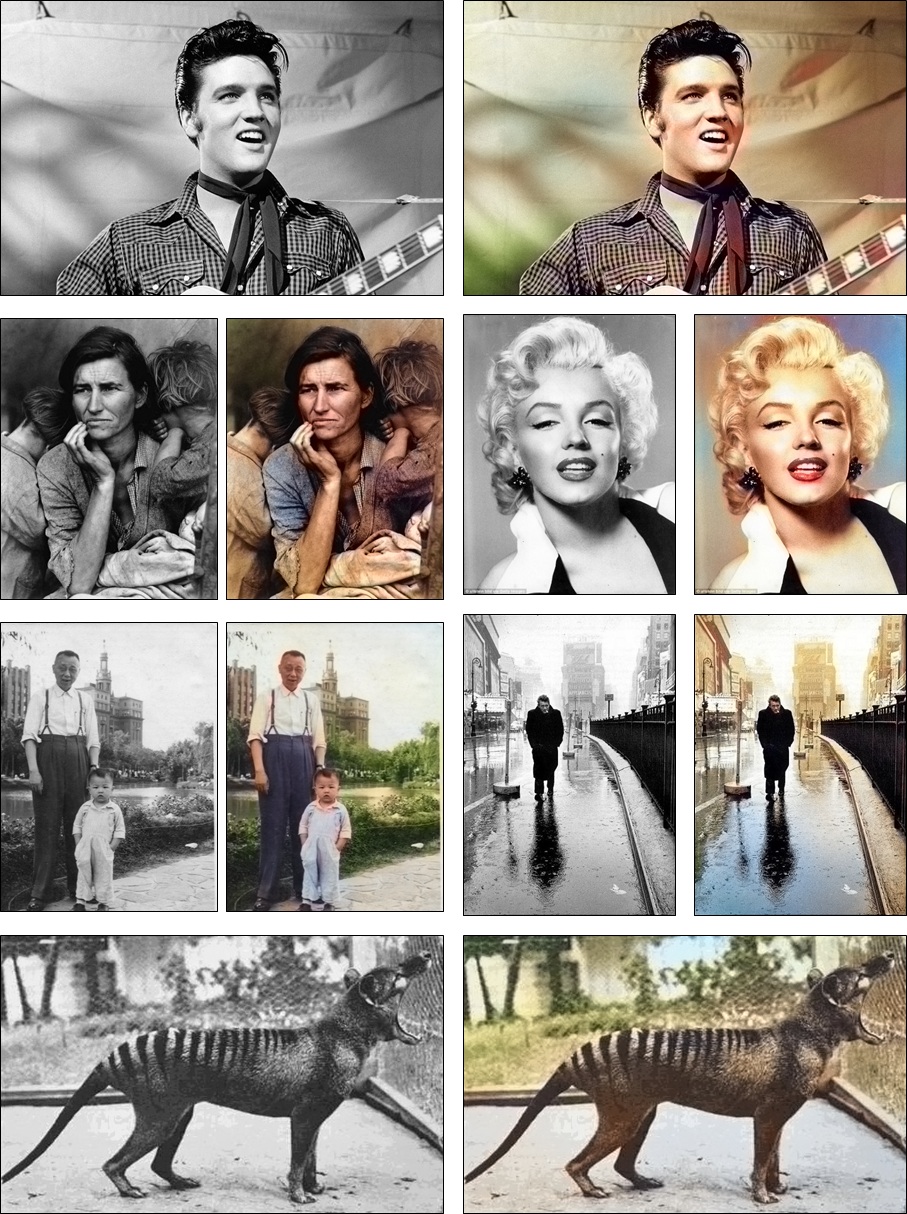}
\caption{Applying our method to legacy black and white photographs. Top to bottom, left to right: photo of Elvis Presley, photo of \textit{Migrant Mother} by Dorothea Lange, photo of Marilyn Monroe, an amateur family photo, photo by Henri Cartier-Bresson, photo by Dr. David Fleay of \textit{Benjamin}, the last captive thylacine which went extinct in 1936.}
 \label{fig:famous_examples}
\end{figure}

Here, we show additional qualitative examples of applying our model to legacy black and white photographs. Figures~\ref{fig:famous_examples_adams}, \ref{fig:famous_examples_bresson}, and \ref{fig:famous_examples} show examples including work of renowned photographers, such as Ansel Adams and Henri Cartier-Bresson, photographs of politicians and celebrities, and old family photos. One can see that our model is often able to produce good colorizations, even though the low-level image statistics of old legacy photographs are quite different from those of modern-day photos. 

\end{document}